\documentclass{article}

\usepackage{iclr2027_conference,times}
\usepackage[utf8]{inputenc}
\usepackage[T1]{fontenc}
\usepackage{hyperref}
\usepackage{url}
\usepackage{booktabs}
\usepackage{amsmath,amssymb,amsfonts,amsthm}
\usepackage{mathtools}
\usepackage{microtype}
\usepackage{xcolor}
\usepackage{graphicx}
\usepackage{enumitem}
\usepackage{multirow}
\usepackage{array,tabularx}
\usepackage{caption}
\usepackage{subcaption}
\usepackage{float}
\usepackage{tikz}
\usepackage{comment}

\usetikzlibrary{arrows.meta,positioning,fit,backgrounds}
\usepackage[capitalize,noabbrev]{cleveref}

\usepackage{iclr2027_arxiv}

\usepackage[textwidth=2.5cm]{todonotes}

\newcommand{\yli}[2][]{\todo[color=red!20, size=\tiny, #1]{\textbf{YL:} #2}}

\newtheorem{theorem}{Theorem}
\newtheorem{proposition}[theorem]{Proposition}
\newtheorem{prop}[theorem]{Proposition}

\theoremstyle{definition}

\theoremstyle{remark}

\newcommand{\pp}{\,\mathrm{pp}}

\crefformat{section}{\S#2#1#3}
\crefformat{subsection}{\S#2#1#3}
\crefformat{subsubsection}{\S#2#1#3}
\crefname{prop}{proposition}{propositions}

\definecolor{linkblue}{HTML}{153E75}
\definecolor{gateblue}{HTML}{2563EB}
\definecolor{gateorange}{HTML}{D97706}
\definecolor{gategreen}{HTML}{059669}
\hypersetup{
    colorlinks=true,
    linkcolor=linkblue,
    citecolor=linkblue,
    urlcolor=linkblue,
    pdftitle={Privileged Likelihood is not Automatically Value}
}

\renewcommand{\arraystretch}{1.08}
\setlist{leftmargin=*,nosep}
\newcolumntype{L}[1]{>{\raggedright\arraybackslash}p{#1}}
\newcolumntype{R}[1]{>{\raggedleft\arraybackslash}p{#1}}

\usepackage[normalem]{ulem}

\title{Privileged Likelihood is not Automatically Value:\\
Three Checks for Token Credit in On-Policy Self-Distillation}
\author{%
Xuan-Phi Nguyen\thanks{Corresponding author:
\texttt{xnguyen@salesforce.com}.},
Shrey Pandit, Zeyu Liu, Yang Li, Yiran Zhao, Anurag Koul, Shafiq Joty\\
Salesforce AI Research
}

\seticlrarxivauthors{%
  Xuan-Phi Nguyen$^{*}$, Zeyu Leo Liu, Yang Li, Shrey Pandit,
  Yiran Zhao, Anurag Koul, Shafiq Joty$^{*}$
}
\seticlrarxivaffiliation{Salesforce AI Research}
\seticlrarxivauthornote{%
  $^{*}$Corresponding author:
  \href{mailto:xnguyen@salesforce.com}{\{xnguyen, sjoty\}@salesforce.com}%
}

\iclrarxivcopy

\ificlrfinal
  \hypersetup{pdfauthor={Xuan-Phi Nguyen, Zeyu Leo Liu, Yang Li, Shrey Pandit, Yiran Zhao, Anurag Koul, Shafiq Joty}}
\else
  \ificlrarxiv
    \hypersetup{pdfauthor={Xuan-Phi Nguyen, Zeyu Leo Liu, Yang Li, Shrey Pandit, Yiran Zhao, Anurag Koul, Shafiq Joty}}
  \else
    \hypersetup{pdfauthor={Anonymous Authors}}
  \fi
\fi

\begin{document}

\maketitle 

\begin{abstract}

On-policy self-distillation aims to improve upon reinforcement learning from verifiable rewards (RLVR) by providing token-level scores derived from privileged information, such as reference solutions or critic feedback available only during training. These scores are implicitly treated as estimates of token-level action values, yet they answer a fundamentally different question: how the model's prediction changes when its input context is enriched, rather than how the expected outcome changes when a token is changed. We examine this gap along three dimensions: (i) whether the token-level score tracks task success; (ii) whether feedback generated from the same rollout causes the score to reflect agreement with its own description, and whether using feedback from other rollouts in the group mitigates this self-referential effect; and (iii) what behavior the resulting training objective actually reinforces. 

In experiments with gpt-oss-20b on AIME~2025, the implemented two-sided score distinguishes correct from incorrect trajectories at approximately chance level (AUC=0.505); using feedback from a different rollout does not consistently improve this discrimination; and all five token-score training configurations achieve only 24.2--33.9\% Avg@4, compared with 64.2\% for outcome-only GRPO. Moreover, the highest-entropy token decile accounts for 57--71\% of the total absolute token-advantage mass, despite the score being least informative about reasoning quality in this regime. By contrast, similar experiments with Qwen3-8B on SciKnowEval Biology improves held-out Avg@8 by 26.8--28.0\%, while its corresponding trajectory scores achieve AUCs of 0.813--0.924. Together, these results suggest that dense credit assignment through distillation can be effective when its likelihood-based scores are empirically validated as meaningful proxies for outcome-relevant credit. When this alignment does not hold, however, the resulting supervision can fail to generalize and may substantially underperform outcome-based RL.

\end{abstract}

\section{Introduction}\label{sec:intro}

Reinforcement learning from verifiable rewards (RLVR) has become a widely used approach for training large language models (LLMs) to reason~\citep{grpo_shao2024deepseekmath}. In this setting, an exact-answer verifier evaluates only the model's final output and returns a single scalar reward for the entire reasoning trajectory. Policy-gradient methods such as Group Relative Policy Optimization (GRPO) then broadcast the resulting trajectory-level advantage to every response token. Although each token contributes differently to the gradient through its log-probability, the verifier itself provides no indication of which reasoning steps were responsible for success or failure. Consequently, a decisive algebraic insight and a routine formatting token receive the same outcome-level supervision. 


On-policy distillation (OPD) provides a natural mechanism for obtaining denser supervision than sequence-level rewards alone~\citep{gkd,minillm}. A snapshot \(\pi_{\mathrm{old}}\) of the student policy \(\pi_\theta\) first generates a rollout \(y\sim\pi_{\mathrm{old}}(\cdot\mid x)\). A teacher then re-evaluates the student's trajectory and produces token-level probability distributions, which we denote by \(q_{\mathrm{score}}\). These distributions provide fine-grained supervision that complements the sparse outcome reward. On-policy \emph{self}-distillation (OPSD)~\citep{opsd,sdpo,rlsd,rlcsd,cripo} is a special case of this framework in which the student and teacher are the same model: \(q_{\mathrm{score}}\) can be obtained from a frozen copy of \(\pi_{\mathrm{old}}\). Crucially, in OPSD, the teacher is conditioned on \emph{privileged information} \(c\), such as a reference solution, environment feedback, or a judge-generated critique, that is available only during training.
The resulting token-level signal, which we term the \emph{privileged-likelihood score}, is defined as
\begin{equation}
\label{eq:score}
d(s_t,v)
=
\log q_{\mathrm{score}}(v\mid s_t,c)
-
\log \pi_{\mathrm{old}}(v\mid s_t),
\end{equation}
where \(s_t=(x,y_{<t})\) denotes the reasoning state. The score compares the privileged scorer's probability for candidate token \(v\) with that of the rollout policy. When \(q_{\mathrm{score}}\) is a copy of \(\pi_{\mathrm{old}}\), this comparison isolates the change induced by \(c\). 
RLCSD~\citep{rlcsd} instead compares a helpful context \(c^+\) with a negative context \(c^-\), yielding the two-sided contrast
\begin{equation}
\label{eq:score2}
d_\pm(s_t,v)
=
\log q_{\mathrm{score}}(v\mid s_t,c^+)
-
\log q_{\mathrm{score}}(v\mid s_t,c^-).
\end{equation}

In both cases, the privileged-likelihood score is implicitly treated as a proxy for the token-level action value \(Q^\pi(s_t,v)\): the expected terminal utility obtained by choosing token \(v\) at state \(s_t\) and then continuing generation under policy \(\pi\). If this approximation is accurate, incorporating privileged-likelihood scores into a sequence-level RL objective can reduce gradient variance by concentrating learning on outcome-critical tokens. If the approximation is poor, however, the resulting supervision introduces bias, encouraging the model to optimize for changes in likelihood that do not correspond to genuine contributions to task success.

Privileged-likelihood scores and action values capture fundamentally different signals. A privileged-likelihood score measures how the model’s preference for a token changes when conditioned on additional privileged information, whereas an action value estimates the expected task outcome after selecting that token and continuing the generation. Their agreement is therefore not guaranteed and must be established empirically. Recent studies indeed show that OPSD can fail in several ways: the teacher may favor adherence to a reference trajectory rather than task correctness, remain insensitive to whether the privileged solution itself is correct, or inadvertently suppress useful exploration~\citep{harne2026privileged,ichihara2026context,kaur2026rethinking,zhu2026manyfaces}.

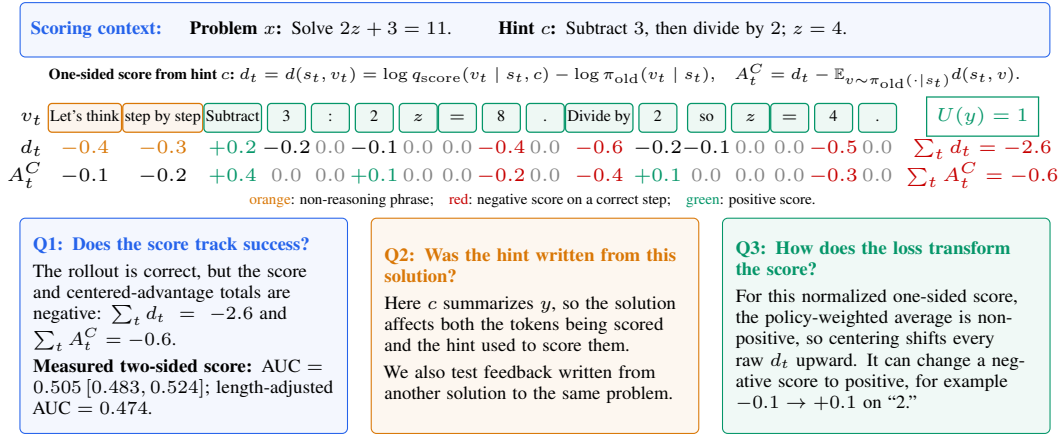
\begin{figure}[t]
\centering
\begingroup
\newcommand{\tokcell}[2]{%
    \tikz[baseline=(token.base)]{
        \node[
            rounded corners=1.2pt,
            draw=#1,
            fill=#1!7,
            minimum width=12pt,
            inner xsep=0.6pt,
            inner ysep=1.8pt,
            font=\fontsize{5.1}{5.5}\selectfont
        ] (token) {\strut #2};
    }%
}
\newcommand{\neutralval}[1]{\textcolor{black!48}{#1}}
\newcommand{\irrelevantval}[1]{\textcolor{gateorange}{\bfseries #1}}
\newcommand{\changedval}[1]{\textcolor{gategreen}{\bfseries #1}}
\newcommand{\wrongval}[1]{\textcolor{red!78!black}{\bfseries #1}}
\begin{tikzpicture}[
    teacherbox/.style={
        rounded corners=2pt,
        align=left,
        inner sep=4pt,
        text width=0.955\linewidth,
        minimum height=0.72cm,
        font=\scriptsize
    }
]
\node[teacherbox,draw=gateblue,fill=gateblue!5] {
    \textbf{\textcolor{gateblue}{Scoring context:}}
    \quad \textbf{Problem \(x\):} Solve \(2z+3=11\).
    \qquad \textbf{Hint \(c\):} Subtract \(3\), then divide by \(2\);
    \(z=4\).
};
\end{tikzpicture}

\vspace{2pt}
{\fontsize{5.7}{6.2}\selectfont
\textbf{One-sided score from hint \(c\):}
\(
d_t=d(s_t,v_t)
    =\log q_{\mathrm{score}}(v_t\mid s_t,c)-\log\pi_{\mathrm{old}}(v_t\mid s_t),
\quad
A^C_t=d_t-\mathbb E_{v\sim\pi_{\mathrm{old}}(\cdot\mid s_t)}d(s_t,v).
\)

\vspace{2pt}
\setlength{\tabcolsep}{0pt}
\renewcommand{\arraystretch}{1.16}
\resizebox{\linewidth}{!}{%
\begin{tabular}{@{}r@{\hspace{1.5pt}}*{17}{c}@{\hspace{3pt}}c@{}}
\(v_t\)
& \tokcell{gateorange}{Let's think}
& \tokcell{gateorange}{step by step}
& \tokcell{gategreen}{Subtract}
& \tokcell{gategreen}{3}
& \tokcell{gategreen}{:}
& \tokcell{gategreen}{2}
& \tokcell{gategreen}{\(z\)}
& \tokcell{gategreen}{\(=\)}
& \tokcell{gategreen}{8}
& \tokcell{gategreen}{.}
& \tokcell{gategreen}{Divide by}
& \tokcell{gategreen}{2}
& \tokcell{gategreen}{so}
& \tokcell{gategreen}{\(z\)}
& \tokcell{gategreen}{\(=\)}
& \tokcell{gategreen}{4}
& \tokcell{gategreen}{.}
& \changedval{\(\boxed{U(y)=1}\)} \\[1pt]
\(d_t\)
& \irrelevantval{\(-0.4\)}
& \irrelevantval{\(-0.3\)}
& \changedval{\(+0.2\)}
& \(-0.2\)
& \neutralval{\(0.0\)}
& \(-0.1\)
& \neutralval{\(0.0\)}
& \neutralval{\(0.0\)}
& \wrongval{\(-0.4\)}
& \neutralval{\(0.0\)}
& \wrongval{\(-0.6\)}
& \(-0.2\)
& \(-0.1\)
& \neutralval{\(0.0\)}
& \neutralval{\(0.0\)}
& \wrongval{\(-0.5\)}
& \neutralval{\(0.0\)}
& \wrongval{\(\sum_t d_t=-2.6\)} \\[1pt]
\(A^C_t\)
& \(-0.1\)
& \(-0.2\)
& \changedval{\(+0.4\)}
& \neutralval{\(0.0\)}
& \neutralval{\(0.0\)}
& \changedval{\(+0.1\)}
& \neutralval{\(0.0\)}
& \neutralval{\(0.0\)}
& \wrongval{\(-0.2\)}
& \neutralval{\(0.0\)}
& \wrongval{\(-0.4\)}
& \changedval{\(+0.1\)}
& \neutralval{\(0.0\)}
& \neutralval{\(0.0\)}
& \neutralval{\(0.0\)}
& \wrongval{\(-0.3\)}
& \neutralval{\(0.0\)}
& \wrongval{\(\sum_t A^C_t=-0.6\)}
\end{tabular}
}

\vspace{1pt}
\textcolor{gateorange}{orange}: non-reasoning phrase;\quad
\textcolor{red!78!black}{red}: negative score on a correct step;\quad
\textcolor{gategreen}{green}: positive score.
}

\vspace{3pt}
\begin{tikzpicture}[
    check/.style={
        rounded corners=2pt,
        align=left,
        inner sep=5pt,
        text width=0.285\linewidth,
        minimum height=2.85cm,
        font=\scriptsize
    }
]
\node[check,draw=gateblue,fill=gateblue!5] (qone) {
    \textbf{\textcolor{gateblue}{Q1: Does the score track success?}}\\[2pt]
    The rollout is correct, but the score and centered-advantage totals are negative:
    \(\sum_t d_t=-2.6\) and \(\sum_t A^C_t=-0.6\).\\[2pt]
    \textbf{Measured two-sided score:}
    AUC \(=0.505\,[0.483,0.524]\); length-adjusted AUC \(=0.474\).
};
\node[check,draw=gateorange,fill=gateorange!6,right=3mm of qone] (qtwo) {
    \textbf{\textcolor{gateorange}{Q2: Was the hint written from this solution?}}\\[2pt]
    Here \(c\) summarizes \(y\), so the solution affects both the tokens being
    scored and the hint used to score them.\\[2pt]
    We also test feedback written from another solution to the same problem.
};
\node[check,draw=gategreen,fill=gategreen!6,right=3mm of qtwo] (qthree) {
    \textbf{\textcolor{gategreen}{Q3: How does the loss transform the score?}}\\[2pt]
    For this normalized one-sided score, the policy-weighted average is
    nonpositive, so centering shifts every raw \(d_t\) upward. It can change
    a negative score to positive, for example \(-0.1\to+0.1\) on ``2.''
};
\end{tikzpicture}
\endgroup
\caption{\textbf{A correct rollout can receive a negative one-sided score.}
The token values shown are illustrative, while Q1 AUC values are measured on AIME~2025. Although the hint \(c\) specifies the correct steps, the conditioned model assigns a lower total likelihood to the observed correct rollout than does the rollout policy. \textbf{Q1} evaluates the score against task success. \textbf{Q2} tests whether the hint was written after observing the same solution. \textbf{Q3} tracks how the policy-weighted baseline shifts the score prior to training.}
\label{fig:token-aligned-failure}
\end{figure}

We examine three questions before treating the privileged-likelihood score as token credit. \textbf{Q1: Does the score track task success?} A likelihood change need not favor a correct solution. \Cref{fig:token-aligned-failure} shows a rollout example with utility \(U(y)=1\) (i.e., correct), but gets negative totals for both \(d_t\) and \(A^C_t\). \textbf{Q2: Was the feedback written from the rollout it scores?} If so, the resulting score may reflect how well the rollout matches feedback written specifically about it, and not whether those tokens contribute to a correct answer. \textbf{Q3: What does the training loss do with the score?} Baseline subtraction, clipping, and token weighting determine which scores become parameter updates.

We examine these questions primarily with \texttt{gpt-oss-20b} trained on DeepScaleR~\citep{deepscaler2025} and evaluated on AIME~2025~\citep{aime2025}. An external LLM judge generates a helpful explanation \(c^+\) and a plausible incorrect explanation \(c^-\) for each rollout. For Q1, we also report Qwen3-8B trained with SDPO~\citep{sdpo} on SciKnowEval Biology~\citep{sciknoweval}, where the privileged information is the gold answer. Our findings are:

\textbf{Q1: The score does not track task outcomes on AIME.} 
On AIME, the implemented two-sided score distinguishes correct from incorrect trajectories only at random chance level (pooled AUC \(=0.505\), 95\% CI \([0.483,0.524]\)); after adjusting for response length, the AUC is \(0.474\). The one-sided score also remains negative for both correct and incorrect rollouts, with only a small relative difference of \(+0.017\). In both cases, the methods underperforms the outcome-only baseline.

When the score does track outcomes, the same check predicts a different training result. On SciKnowEval Biology, Qwen3-8B trained with SDPO~\citep{sdpo} using the gold answer as privileged information achieves trajectory-level AUC \(0.813\)--\(0.924\) and improves held-out Avg@8 by 26.8--28.0\%. Outcome separation is therefore a practical first check of the score.


\textbf{Q2: Feedback from another rollout does not recover a useful score.}
When feedback is written after reading the rollout being scored, the score may reflect textual agreement with that particular feedback. We therefore score each rollout with feedback written from a different rollout of the same problem, a procedure called \emph{cross-fitting}. The resulting AUCs remain near chance or change direction across comparisons on AIME, so changing which rollout supplies the feedback does not make the score reliably track correctness.


\textbf{Q3: Different token-level scoring configurations consistently degrade training.}
We explore five different ways to integrate the token-level score into the advantage function during RL training. Particularly, we evaluate three additive configurations and two KL-regularized configurations against outcome-only GRPO. The outcome-only baseline achieves 64.2\% Avg@4, whereas the token-score configurations achieve 24.2\%--33.9\%, a decrease of 30--40\%. A separate analysis finds that 57--71\% of the sum of absolute token advantages falls in the highest-entropy token decile, indicating that the learning objective places most token-level weight at positions where the model is most uncertain.
Together, the experiments show that the score, the feedback used to compute it, and the training loss must be checked separately before a likelihood difference should be treated as token credit.
Our contributions can be summarized as:
\begin{itemize}
\item We organize privileged self-distillation around three diagnostic questions---whether token scores track task success (Q1), whether rollouts are compared under a controlled scoring rule (Q2), and what behavior the training loss reinforces (Q3)---and provide theoretical analyses and proofs to support the relationships between these questions and outcome success.

\item On AIME 2025 with gpt-oss-20b reasoning model, we find that the implemented score separates correct and incorrect trajectories at random chance level. Meanwhile, on SciKnowEval Biology, we find a positive setting in which stronger outcome separation accompanies successful SDPO training, supporting Q1 as a practical screen.

\item Cross-fitting with feedback from another rollout does not consistently improve performance, consistent with previous studies~\citep{ichihara2026context}.

\item All tested token-score configurations underperform outcome-only GRPO, indicating that adding these token scores does not improve training in this setting.


\end{itemize}

\section{Background and Related Work}\label{sec:background}

\subsection{The OPSD recipe and its variants}

The one-sided score \(d(s_t,v)\) and two-sided score \(d_\pm(s_t,v)\) defined in Equations \ref{eq:score} and \ref{eq:score2}, respectively, supply token-level signals, while the training loss determines how those signals update the model. Existing methods use them in three different ways.

\textbf{As a distillation target.}
The simplest use treats the teacher distribution \(q_{\mathrm{score}}(\cdot\mid s_t,c)\) as a soft label and minimizes a divergence between it and the student \(\pi_\theta(\cdot\mid s_t)\). OPSD~\citep{opsd} and SDPO~\citep{sdpo} apply this distillation across the rollout via a KL divergence objective, while CriPO~\citep{cripo} restricts it to tokens selected by a rubric. VPD~\citep{vpd} co-evolves the teacher from language feedback before this distillation step. These losses match the teacher distribution without using task reward. The frozen-target projection is derived in \cref{prop:fkl-projection}.

\textbf{As an additive reward.}
A second family combines the token score with a verifier reward. RLSD~\citep{rlsd} uses the score to weight each token's contribution to the utility advantage. RLCSD~\citep{rlcsd} compares helpful and incorrect contexts while retaining the verifier reward. OPPO~\citep{oppo} derives a Bayesian recursion that accumulates the log-ratio along a trajectory to estimate running success probability. SC-GRPO~\citep{scgrpo} uses the per-token KL from a self-conditioned teacher as a multiplicative weight on the utility (outcome) advantage. These methods are \emph{reward-anchored}: the token score modulates an objective that already includes verifier feedback, and when the token score is zeroed out, they reduce to a standard outcome-only method.

\textbf{As a KL regularizer.}
A third group uses the privileged teacher as a KL target and adds a penalty \(\beta D_{\mathrm{KL}}(\pi_\theta(\cdot\mid s_t)\|q_{\mathrm{score}}(\cdot\mid s_t,c))\) to the objective. This resembles the form of KL-regularized control~\citep{regularizedmdp,dpo}. VPD~\citep{vpd} derives its student reverse-KL update from this optimum. The exact optimum is an exponential tilt of a reference policy by a soft advantage (\cref{prop:gibbs-tether}). But this method requires that the target remain fixed across optimization and that the KL penalty be the only shaping term; on-policy self-distillation may refresh \(q_{\mathrm{score}}\) as \(\pi_\theta\) changes across updates.

A noisy score may be harmful as a standalone distillation target but useful as a small addition to an outcome reward, while a systematically incorrect score can harm either objective. This is why the training loss (Q3 in our framework) must be diagnosed separately from the score itself (Q1).

\subsection{Concurrent empirical landscape}

The privileged self-distillation literature has grown rapidly recently. We summarize the concurrent findings most relevant to our analysis.

\textbf{Failure reports.}
\citet{harne2026privileged} study OPSD and find that the teacher follows the reference trajectory more closely than correctness, with much of the supervision falling on inconsequential tokens such as stopwords and punctuation, and with a per-token loss that is indistinguishable between correct and incorrect rollouts. \citet{ichihara2026context}, meanwhile, find that a worked solution, even a deliberately corrupted one, from another problem can retain much of the OPSD gain, showing that the problem's own solution is not always necessary and that whatever the teacher transfers is not target-specific correctness. \citet{kaur2026rethinking} report that privileged teachers can suppress exploratory tokens at uncertain reasoning steps, while \citet{zhu2026manyfaces} report better results with shared rules than with sample-specific solutions.

\textbf{Success reports.}
A parallel literature reports gains when the token score is combined with a verifier anchor or when the privileged information is constructed differently. OPPO~\citep{oppo} derives a Bayesian credit recursion anchored to the GRPO advantage and reports improvements on Math. Purified OPSD~\citep{shen2026purified} decomposes the teacher update along the question-presence axis---subtracting a reference-only teacher (no question) from the full teacher (question plus reference), making this an orthogonal decomposition to our correctness-polarity contrast. SC-GRPO~\citep{scgrpo} uses per-token self-conditioned KL as a multiplicative weight on a verifier-anchored objective, while CAST~\citep{cast} uses answer-free self-teacher gaps to shape GRPO advantages. VPD~\citep{vpd} co-evolves a feedback-conditioned teacher and reports gains over RLVR and frozen-teacher self-distillation on scientific reasoning and code generation. PMD~\citep{pmd} conditions the teacher on online procedural memory and improves over SDPO. Several other methods---position-weighted scoring~\citep{pwopsd}, local error correction~\citep{rosd}, sibling-conditioned distillation~\citep{ssopd}---modify where or how the score is applied while retaining the basic privileged-likelihood construction. $\beta$-OPSD~\citep{betaopsd} addresses the myopia of per-token KL updates with a return-to-go estimator that recovers an unbiased sequence-level gradient.


\textbf{The emerging pattern.}
Overall, these results show that OPSD depends on both the information given to the teacher and the way its probabilities enter the loss. Our experiments complement this work by checking whether the score separates correct and incorrect outcomes, whether the feedback written from the rollout being scored or cross-fitting can have any impact, and how different token-level advantage configurations compare with outcome-only training. We also conduct theoretical analyses to establish the links between these conditions to sequence-level outcome success.

\section{Framework}\label{sec:framework}

Before treating a privileged-likelihood score as token credit, we ask three questions:
\begin{enumerate}[leftmargin=2.0em,labelindent=0em,labelsep=0.5em,topsep=2pt,itemsep=1pt,style=sameline]
\item[\textbf{Q1.}] \emph{Does the score track task success?} At a given reasoning state, does the score prefer next tokens that lead to better outcomes?
\item[\textbf{Q2.}] \emph{Was the feedback written from the rollout being scored?} If so, the score may partly measure agreement with a description written specifically for that rollout. Does the result persist when each rollout is scored with feedback from another rollout, when self-dependence is removed?
\item[\textbf{Q3.}] \emph{What does the training loss do with the score?} Baseline subtraction, clipping, and token weighting determine how the score changes the update. Does the resulting gradient point in a helpful direction?
\end{enumerate}
These are separate questions: a score can fail Q1 while Q2 is irrelevant, or pass Q1 while the loss (Q3) destroys the signal. This section establishes why each separation is formally necessary.

\subsection{Q1: When does a likelihood ratio rank action value?}\label{sec:when-rank}

At a reasoning state \(s_t\), let \(d(s_t,v)\) denote the likelihood score being evaluated and let \(Q^\pi(s_t,v)\) be the expected terminal utility (i.e. outcome reward) after choosing \(v\) and continuing with policy \(\pi\). For the two quantities to value every local redistribution of next-token probability in the same relative scale, the score must be a positive affine transformation of the action value. Otherwise, a score that preserves the ranking but distorts the magnitudes will over-concentrate mass on the top-ranked token. For example, action values \(Q=(0,1,2)\) and scores \(d=(0,1,4)\) rank three tokens identically but give different relative weight to probability shifts.

\begin{proposition}[Exact local affine equivalence]\label{prop:affine}
At an interior policy, a score \(d(s,\cdot)\) assigns a proportional first-order change to \(Q^\pi(s,\cdot)\), with one positive scale shared across every zero-sum perturbation of the next-token distribution, if and only if there exist a state-dependent offset \(\omega_0(s)\in\mathbb R\) and scale \(\omega_1(s)>0\), both independent of \(v\), such that
\begin{equation}\label{eq:affine}
d(s,v)=\omega_0(s)+\omega_1(s)\,Q^\pi(s,v)
\quad \text{for every }v.
\end{equation}
\end{proposition}

This is a standard expected-utility result~\citep{hersteinmilnor,rewardinvariance}; the proof appears in \cref{app:affine-proof}.

The reason privileged-likelihood scores are plausible as credit is that a log-ratio of this form does rank action value in one standard case. Hindsight credit assignment~\citep{hca,cocoa} uses the true success and failure conditionals: for each candidate token \(v\), let \(q_1(v\mid s)\) be how likely \(v\) is given that the trajectory will succeed, and \(q_0(v\mid s)\) given that it will fail. Bayes' rule then gives
\begin{equation}\label{eq:bayes-odds}
\log \frac{q_1(v \mid s)}{q_0(v \mid s)} \;=\; \operatorname{logit} P(U\!=\!1 \mid s, v) \;-\; \operatorname{logit} P(U\!=\!1 \mid s),
\end{equation}
where \(P(U=1\mid s,v)=Q^\pi(s,v)\), under the binary reward assumptions stated in \cref{app:outcome-conditionals}. Because the logit is increasing, this log-ratio ranks binary action value. Tokens that lead to success get higher scores.

This identity explains why privileged-likelihood scores are appealing. The two-sided score \(d_\pm(s,v)=\log q_{\mathrm{score}}(v\mid s,c^+)-\log q_{\mathrm{score}}(v\mid s,c^-)\) has the same log-ratio form. The ranking guarantee applies only if the two distributions $q_1$ and $q_0$, expressed via context $c^+$ and $c^-$, are coherent success and failure conditionals from one joint distribution. However, free-form texts labeled ``helpful'' and ``unhelpful'' need not satisfy that condition, and may also differ in formatting, length, or which values they reveal.

This is the gap that Q1 targets. The form of the score guarantees nothing; agreement with action value is an empirical question that must be checked. The most accessible test is at the sequence level: average the token scores over a trajectory and ask whether the resulting scalar separates correct from incorrect rollouts (AUC). This is a necessary but not sufficient condition---if the mean token score cannot even tell correct trajectories from incorrect ones, the individual token scores cannot be ranking action values correctly, since any correct ranking must aggregate to a correct trajectory-level ordering.

\subsection{Q2: Self-dependence and what cross-fitting does and does not fix}\label{sec:cross-fitting}

Q2 concerns the relationship between the privileged context and the rollout being scored. Privileged information comes in many forms, and the forms differ in a structural way that matters for the score's validity. A reference solution or a fixed worked example is determined by the problem alone: it exists before the student generates anything. A judge-written critique or environment feedback, by contrast, is typically generated \emph{after reading} the student's rollout. In the first case the scoring context is independent of the rollout; in the second, it is not.

When the context depends on the rollout, the score picks up an additional source of variation: textual agreement between the rollout and its own description. A critique that says ``do not divide by 4'' is informative precisely because it describes the error the student made---but this means the score depends on \(y\) through both the tokens being scored and the context \(c(y)\). The score can be high simply because the rollout and its feedback are about the same thing, regardless of whether the rollout is correct.

Cross-fitting removes this loop by scoring each rollout with feedback written from a \emph{different} rollout of the same problem. The target rollout no longer influences its own scoring context. This is a clean test of whether the score's apparent signal survives when \(c\) is not written from the target, removing self-dependence.

\begin{proposition}[Target-side independence under cross-fitting]\label{prop:crossfit}
Let \(c\) be feedback constructed only from a donor rollout \(y^{\mathrm{don}}\) and external randomness, and let \(y^{\mathrm{tar}}\) be an independently selected target rollout for the same prompt \(x\) and fold assignment \(f\). Then \(c\perp y^{\mathrm{tar}}\mid x,f\). This independence does not imply that the resulting score tracks task utility.
\end{proposition}

Cross-fitting therefore defines a held-out-feedback score rather than a corrected version of same-rollout scoring. It answers how the score changes when the target does not supply its own feedback. Removing the dependence of \(c\) on the target rollout changes the scoring rule; it does not recover a debiased version of same-rollout scoring. Q1 still determines whether the new score tracks outcomes. The proof and the corresponding partial-gradient analyses are elaborated in \cref{app:crossfit-proof}.

\subsection{Q3: How does the loss turn a score into an update?}\label{sec:score-loss-separation}

Q1 asks what information the score carries. Q3 follows what happens after that score enters training:
\[
d_\pm(s,v)\ \longrightarrow\ A^C(s,v)\ \longrightarrow\ A_{i,t}\ \longrightarrow\ \nabla_{\mathbf z}\mathcal L\ \longrightarrow\ \nabla_\theta\mathcal L.
\]
Here \(A^C\) is the centered token advantage, \(A_{i,t}\) is the final token advantage after combining the score with utility advantage \(A_i^U\) (if any) and applying any gates or clipping, $(\mathcal L)$ is the actor loss, $\mathbf z$ denotes the token-logit vector, and $(\theta)$ denotes the shared model parameters. The following steps distinguish changes to a sampled advantage, the local response of a loss, and the resulting parameter gradient.

\paragraph{From scores to token advantages.}
Given a sampled token \(v\sim\pi_{\mathrm{old}}(\cdot\mid s)\), the policy-weighted baseline gives
\begin{equation}\label{eq:same-slot-baseline}
A^C(s,v)=d_\pm(s,v)-\sum_{v'}\pi_{\mathrm{old}}(v'\mid s)d_\pm(s,v').
\end{equation}
For a two-sided score, the policy-weighted mean can have either sign, so centering can turn a positive score into a negative token advantage. 
There fore, the loss acts on the centered signal rather than the raw score.

\paragraph{Losses respond differently at confident predictions.}
The same score can enter training as a bounded additive term or through KL regularization. For fixed \(d_\pm(s,\cdot)\), reference policy \(\pi_{\mathrm{ref}}\), and positive \(\alpha,\beta\), maximizing \(\alpha\mathbb E_{v\sim\pi_\theta(\cdot\mid s)}d_\pm(s,v)-\beta D_{\mathrm{KL}}(\pi_\theta(\cdot\mid s)\|\pi_{\mathrm{ref}}(\cdot\mid s))\) is equivalent to minimizing reverse KL toward the score-tilted target
\[
\pi^\star(v\mid s)\ \propto\ \pi_{\mathrm{ref}}(v\mid s)\exp\!\left(\frac{\alpha}{\beta}d_\pm(s,v)\right).
\]
This connection permits a common local comparison of additive and KL objectives.

\begin{proposition}[Local response near a confident prediction]\label{prop:boundary-geometry}
Restrict the next-token distribution to two tokens, let \(v\) be the more likely token, and write \(\varepsilon=1-\pi_\theta(v\mid s)\) for the probability of the other token. For a bounded fixed score \(d_\pm(s,\cdot)\), the logit-gradient magnitude of the additive loss \(-\mathbb E_{v'\sim\pi_\theta(\cdot\mid s)}d_\pm(s,v')\) is at most a constant times \(\varepsilon\). For reverse KL \(D_{\mathrm{KL}}(\pi_\theta(\cdot\mid s)\|\pi^\star(\cdot\mid s))\) toward a fixed full-support target \(\pi^\star\), the magnitude is proportional to \(\varepsilon\log(1/\varepsilon)\) near \(\varepsilon=0\). Both gradients approach zero as the model becomes certain.
\end{proposition}

Reverse KL therefore responds more strongly than a bounded additive score near this boundary, but both responses weaken as the policy becomes certain. Their gradients move toward the supplied score or target, which need not favor higher task utility. Full derivations, comparisons with other losses, and saved-state measurements appear in \cref{app:geometry-proof,fig:gradient-geometry}.

\paragraph{From token advantages to parameter gradients.}
Let \(T_i\) be the response length, let \(A_{i,t}\) be the detached advantage assigned to token \(t\), and let \(h_{i,t}=\nabla_\theta\log\pi_\theta(v_{i,t}\mid s_{i,t})\). Let \(g_A\) be the expected ascent update using \(A_{i,t}\), and \(g_U\) for the outcome-only ascent update that uses \(A_i^U\) at every token; the corresponding actor-loss gradients have the opposite sign. Before policy-ratio clipping,
\begin{equation}\label{eq:advantage-gradient-difference}
g_A-g_U
=
\mathbb E\!\left[
\frac{1}{T_i}\sum_t
\left(A_{i,t}-A_i^U\right)h_{i,t}
\right].
\end{equation}
The advantage values alone do not determine this vector because the token score-function vectors \(h_{i,t}\) generally point in different parameter-space directions.

\begin{proposition}[Advantage totals do not determine the parameter gradient]\label{prop:update-nonconservation}
Even if \(A_{i,t}=A_i^Uw_{i,t}\) with positive, mean-one weights \(w_{i,t}\), preserving the advantage sign and its total across tokens does not generally preserve the expected policy-gradient estimate.
\end{proposition}


For two independent token coordinates, changing the weights from \((1,1)\) to \((2-\varepsilon,\varepsilon)\) changes the expected update from \((1/8,1/8)\) to \(((2-\varepsilon)/8,\varepsilon/8)\), although the weights remain positive and mean one. More detailed derivation is in \cref{app:update-counterexample}. Token weighting can be useful precisely because it changes the update; the proposition says that the advantage sign and total across tokens alone cannot certify that change.

\section{Experimental Setup}\label{sec:setup}

We study how privileged-likelihood scores behave in two settings. The main AIME experiments analyze the likelihood score relationships, feedback construction, and different ways of using the score during RL training. Plus, a SciKnowEval Biology experiment provides a positive comparison for the outcome-separation check.

\paragraph{Model and benchmark.}
The AIME study trains \texttt{gpt-oss-20b} with effort \texttt{low} on DeepScaleR \citep{deepscaler2025} and evaluates on AIME~2025~\citep{aime2025}. For each problem, the model samples \(K>1\) solutions. An equivalent-answer verifier is employed to assign binary outcome utility \(U_i\in\{0,1\}\).

\paragraph{Feedback construction.}
For the AIME experiments, an external LLM judge (gpt-5.6-luna) receives the problem, reference solution, verifier result, and full reasoning trace. It produces a helpful feedback \(c_i^+\) and a plausible incorrect or harmful feedback $(c_i^-)$. To prevent the student from merely copying the feedback, the helpful template masks selected numerical values and often includes formatting guidance, while the incorrect template supplies explicit intentionally plausible-yet-incorrect values. The two texts are intended to differ only in the mathematical claim being tested, so that their likelihood contrast isolates mathematical content. The resulting likelihood contrast can therefore respond to wording and formatting as well as mathematical content. The complete feedback and scorer definitions are reported in \cref{app:score-provenance}.


\paragraph{Scoring.}
The rollout policy \(\pi_{\mathrm{old}}\) generates the sampled trajectories. A teacher distribution \(q_{\mathrm{score}}\) supplies the privileged probabilities. The scorer checkpoint used by each analysis is identified in \cref{app:score-provenance}. We use the score notation introduced in \cref{sec:intro}:
\begin{align}
d(s_{i,t},v)
&=\log q_{\mathrm{score}}(v\mid s_{i,t},c_i^+)-\log\pi_{\mathrm{old}}(v\mid s_{i,t}), \\
d_\pm(s_{i,t},v)
&=\log q_{\mathrm{score}}(v\mid s_{i,t},c_i^+)-\log q_{\mathrm{score}}(v\mid s_{i,t},c_i^-).
\label{eq:experimental-scores}
\end{align}
The one-sided score \(d\) supports the teacher--policy comparison in \cref{sec:q1}; the training configurations use the two-sided score $(d_\pm)$. For any token score $f$, let its policy-centered value as
\begin{equation}\label{eq:ctr-value}
\operatorname{ctr}_{s}[f](v)
=f(s,v)-\sum_{v'}\pi_{\mathrm{old}}(v'\mid s)f(s,v'). 
\hspace{1em } A^C_{i,t}=\operatorname{ctr}_{s_{i,t}}[d_\pm](v_{i,t})
\end{equation}

\paragraph{Training configurations.}
The AIME configurations share the initialization, training data, optimizer, rollout batches, and verifier. The outcome-only configuration uses the leave-one-out utility advantage
\begin{equation}\label{eq:loo-outcome}
A_i^U=U_i-\frac{1}{K-1}\sum_{j\neq i}U_j.
\end{equation}
For dense-reward configurations with additive advantage, we applied several technique to stabilize training. We define \(d_\pm^{\mathrm{clip}}=\operatorname{clip}(d_\pm,-2,2)\) and \(A^{C,\mathrm{clip}}_{i,t}=\operatorname{ctr}_{s_{i,t}}[d_\pm^{\mathrm{clip}}](v_{i,t})\). All three additive variants use \(B(z)=0.5\tanh(z/0.5)\), which smoothly bounds each token term to \([-0.5,0.5]\). Source clipping limits extreme log-ratios before centering, while \(g_{H,i,t}=\min(1,H_{i,t}/0.2)\) attenuates the token-level advantage where \(\pi_{\mathrm{old}}(\cdot\mid s_{i,t})\) has low entropy \(H_{i,t}\). The three configurations apply entropy weighting, source clipping, or both.

The KL-regularized variants test whether a fixed reference-policy penalty can limit drift induced by the token score. The full variant uses the entire reference-policy correction, while the projected variant keeps only the component aligned with the score direction. To define these terms, let \(\ell(s,v)=\log\!\frac{\pi_{\mathrm{old}}(v\mid s)}{\pi_{\mathrm{ref}}(v\mid s)}\), where \(\pi_{\mathrm{ref}}\) is fixed. With the implemented scales,
\begin{equation}g
A^{C,\mathrm{full}}_{i,t}=\operatorname{ctr}_{s_{i,t}}[d_\pm^{\mathrm{clip}}-\ell](v_{i,t}),
\qquad
A^{C,\mathrm{proj}}_{i,t}=0.25(1-\widehat\gamma_{s_{i,t}})A^{C,\mathrm{clip}}_{i,t}.
\end{equation}
where \(\widehat\gamma_s\) is the least-squares projection slope in \cref{eq:projected-kl-slope}. In all five, the token-level term is disabled when \(A_i^U=0\). 
We specify the formulas used in the comparison in \cref{tab:fixed-horizon}, while the exact execution order and hyperparameters are in \cref{app:training-rule-configs}.

\paragraph{Cross-fitting protocol.}
For Q2, each rollout is first paired with feedback written from that rollout. We then assign the rollout--feedback pairs within each prompt to folds using a content-independent permutation. Each target rollout is scored using feedback paired with a donor rollout from the opposite fold, so the target does not supply its own scoring context (\cref{sec:cross-fitting}). The resulting AUCs test whether the score's apparent signal survives removal of direct target-to-feedback dependence.

\paragraph{SciKnowEval Biology comparison.}
We also train Qwen3-8B (both thinking ON and OFF) with SDPO on SciKnowEval Biology questions. The teacher's only privileged information is the gold answer, written as ``The correct answer is [gold answer].''

\paragraph{Evaluation.}
For AIME, we report Avg@4 at the latest checkpoint shared by all six configurations, together with paired prompt-bootstrap confidence intervals and exact sign-flip \(p\)-values. For SciKnowEval Biology, we report held-out Avg@8 and trajectory-score AUC. Further scorer, checkpoint-selection, and statistical details are in \cref{app:protocol}.

\section{Results}\label{sec:results}

We evaluate the three diagnostic questions on the system described in \cref{sec:setup}: whether the score tracks correctness (Q1), whether feedback construction changes the comparison (Q2), and what behavior the training loss produces (Q3).

\subsection{Q1: The dense-reward score does not reliably discriminate}\label{sec:q1}

\paragraph{One-sided teacher--policy residual.}
We first examine one-sided $d(s_{i,t},v)$, by measuring the mean residual $\frac{1}{T}\sum_{t=1}^T d(s_{i,t},v)$ across 18 reasoning checkpoints trained on DeepScaleR.
If this residual provides useful credit, it should be positive for tokens from correct rollouts (the teacher supports the student's choices) and negative for tokens from incorrect ones (the teacher suppresses them).
The empirical mean residual, however, is negative for both correct and incorrect rollouts (\(-0.0334\) and \(-0.0502\)).
The teacher down-scores the student's own sampled tokens regardless of correctness. The gap between correct and incorrect ($+0.017$) is positive, indicating that the privileged context retains some relative outcome information, but the absolute signal points in the wrong direction for both.

\paragraph{Two-sided trajectory AUC.}
We next test the two-sided score \(d_\pm\) used in training. For each trajectory, we average the bounded, centered token advantages over response tokens and measure whether the result ranks a correct trajectory above an incorrect one. We measure that with AUC, which is the probability that a randomly drawn correct trajectory scores above a randomly drawn incorrect one (0.5 is chance).
Pooled AUC compares all trajectories, length-adjusted AUC removes the score's linear association with response length, and within-prompt AUC compares solutions to the same problem.

\begin{table}[H]
\centering
\caption{\textbf{Correctness discrimination on AIME 2025.} Values are AUCs with 95\% prompt-bootstrap confidence intervals. Each row uses its predefined evaluation window. 
}
\label{tab:score-auc}
\scriptsize
\setlength{\tabcolsep}{2.6pt}
\begin{tabular}{@{}lccc@{}}
\toprule
Score construction & Pooled AUC & Length-adjusted AUC & Within-prompt AUC \\
\midrule
Implemented two-sided score
  & \(0.505\;[0.483,\,0.524]\)
  & \(0.474\;[0.452,\,0.494]\)
  & \(0.499\;[0.469,\,0.533]\) \\
Feedback from the same rollout
  & \(0.588\;[0.500,\,0.673]\)
  & \(0.551\;[0.463,\,0.625]\)
  & \(0.609\;[0.542,\,0.670]\) \\
Other-rollout masked feedback
  & \(0.512\;[0.440,\,0.579]\)
  & \(0.485\;[0.423,\,0.548]\)
  & \(0.516\;[0.450,\,0.574]\) \\
Other-rollout correct-fact feedback
  & \(0.416\;[0.347,\,0.484]\)
  & \(0.446\;[0.382,\,0.513]\)
  & \(0.587\;[0.525,\,0.646]\) \\
\bottomrule
\end{tabular}
\end{table}

As shown in \Cref{tab:score-auc}, the implemented score is indistinguishable from random chance in the pooled comparison (AUC = 0.505) and slightly favors incorrect trajectories after length adjustment (AUC = 0.474).
This answers Q1 negatively for the AIME experiment: the implemented training signal does not reliably rank correct trajectories above incorrect ones.
Because any score that provides accurate token-level credit must at minimum separate trajectories in aggregate, this sequence-level failure already implies failure at the token level. The near-chance AUC does not mean the available information is useless. An external judge given the same inputs selected the better solution at $\sim 94\%$ accuracy, yet the base model's token-likelihood score does not extract that information.


\paragraph{Positive comparison on SciKnowEval Biology.}
The same diagnostic behaves differently when Qwen3-8B is trained by SDPO with the gold answer as privileged information. Both thinking configurations improve held-out Avg@8, and their raw and centered trajectory scores remain highly above chance.

\begin{table}[H]
\centering
\caption{\textbf{One-sided SDPO with gold-answer context on SciKnowEval Biology.} Avg@8 accuracy on 50 held-out questions. Within-prompt AUC uses the raw trajectory-mean score.}
\label{tab:sciknow-positive}
\scriptsize
\setlength{\tabcolsep}{3.2pt}
\begin{tabular}{@{}lccccc@{}}
\toprule
Configuration & Avg@8, step \(0\to100\) & Gain (\(\pp\)) & Raw AUC & Centered AUC & Within-prompt AUC \\
\midrule
Thinking disabled & \(31.50\to58.25\) & \(+26.75\) & \(0.924\) & \(0.751\) & \(0.833\) \\
Thinking enabled  & \(26.25\to54.25\) & \(+28.00\) & \(0.813\) & \(0.626\) & \(0.709\) \\
\bottomrule
\end{tabular}
\end{table}

Centering reduces discrimination in both configurations but does not remove it. Across these two systems, stronger outcome separation accompanies successful self-distillation training, supporting trajectory AUC as a practical Q1 screen.

\subsection{Q2: Feedback from another rollout does not recover a reliable score}\label{sec:q2}

The same-rollout score in \cref{tab:score-auc} reaches pooled AUC \(0.588\), but each solution helped determine the feedback used to score it. Cross-fitting instead uses feedback written from another rollout of the same problem. The two other-rollout scores have pooled AUCs \(0.512\) and \(0.416\), and the latter changes to \(0.587\) within prompt. Changing the feedback source therefore changes what the score measures, but neither held-out construction gives a stable relationship with correctness across the reported comparisons.
This answers Q2: removing self-dependence is mechanically sound but does not rescue the score. The concurrent finding of \citet{ichihara2026context} that replacing the target's reference solution with a different problem's solution produces comparable training outcomes. Thus, whatever the privileged context transfers, it is not target-specific credit.

\subsection{Q3: All tested token-score configurations underperform}\label{sec:q3}

We compare outcome-only GRPO and five token-level dense-reward configurations on the same AIME~2025 evaluation set at the latest shared checkpoint.

\begin{table}[H]
\centering
\caption{\textbf{Paired AIME 2025 comparison.} Accuracy is Avg@4. Differences are computed before rounding as configuration minus outcome-only GRPO, with 95\% paired prompt-bootstrap confidence intervals; negative values favor outcome-only GRPO. 
More details can be found in \cref{app:training-rule-configs}.
}
\label{tab:fixed-horizon}
\footnotesize
\setlength{\tabcolsep}{2pt}
\begin{tabular*}{\linewidth}{@{\extracolsep{\fill}}L{0.32\linewidth}lcc@{}}
\toprule
Configuration & Core advantage & Acc. (Avg@4) & Difference [95\% CI] (\(\pp\)) \\
\midrule
Outcome-only GRPO
  & \(A_i^U\)
  & \(64.2\) & -- \\
Entropy-gated additive
  & \(A_i^U+g_{H,i,t}B(A^C_{i,t})\)
  & \(33.9\) & \(-30.4\;[-42.9,\,-17.9]\) \\
Source-clipped additive
  & \(A_i^U+B(A^{C,\mathrm{clip}}_{i,t})\)
  & \(27.5\) & \(-36.7\;[-49.2,\,-24.2]\) \\
Entropy-gated + source-clipped
  & \(A_i^U+g_{H,i,t}B(A^{C,\mathrm{clip}}_{i,t})\)
  & \(30.0\) & \(-34.2\;[-45.8,\,-23.3]\) \\
Full KL-regularized
  & \(A_i^U+A^{C,\mathrm{full}}_{i,t}\)
  & \(24.2\) & \(-40.0\;[-52.5,\,-27.5]\) \\
Projected KL correction
  & \(A_i^U+A^{C,\mathrm{proj}}_{i,t}\)
  & \(25.0\) & \(-39.2\;[-50.8,\,-27.5]\) \\
\bottomrule
\end{tabular*}
\end{table}

All five dense-reward configurations underperform outcome-only GRPO. Outcome-only GRPO reaches \(64.2\%\) accuracy, while the tested dense variants range from \(24.2\%\) to \(33.9\%\), corresponding to decreases of \(30.4\) to \(40.0\) percentage points. Every confidence interval is entirely below zero.

\begin{figure}[H]
\centering
\includegraphics[width=0.94\linewidth]{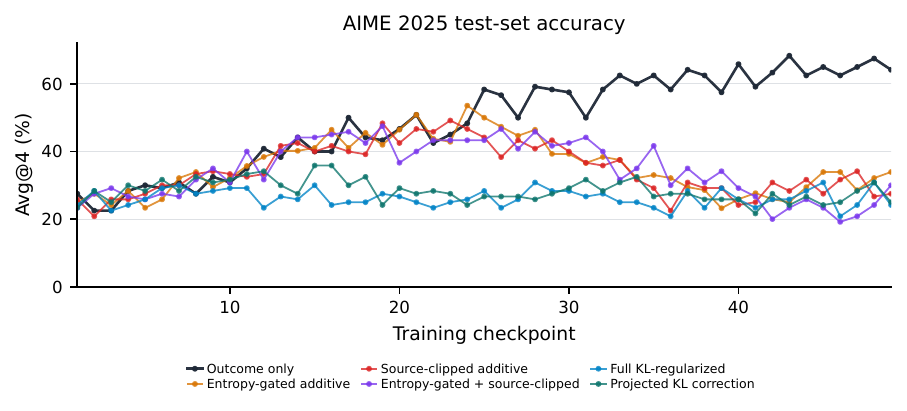}
\caption{\textbf{AIME 2025 accuracy over training.} Test Avg@4 begins in a similar range across configurations. Outcome-only GRPO continues to improve, while all five token-score configurations finish below it. The fixed comparison in \cref{tab:fixed-horizon} uses update~49.}
\label{fig:checkpoint-trajectories}
\end{figure}

\Cref{fig:checkpoint-trajectories} shows how the performance gap develops during training. Outcome-only GRPO improves and stabilizes, while the token-level dense-reward configurations plateau or decline. The dense-reward configurations also produce boxed, complete answers more often than outcome-only GRPO, so their lower accuracy is not explained by a greater failure to return final answers.

\paragraph{How the score changes training.}
The five paired configurations disable the token advantage when \(A_i^U=0\). In a separate ungated additive analysis, \(57\%\)--\(71\%\) of the sum of absolute token advantages comes from the highest-entropy token decile. Across the paired configurations, the additive variants move toward higher token entropy and larger reference-policy KL, while the KL-regularized variants keep both quantities near the outcome-only range but gain no accuracy, as shown in \Cref{fig:checkpoint-trajectories}.

Centering can also change the sign of an individual token advantage, as illustrated in \cref{fig:token-aligned-failure}. The implemented loss then combines the centered score with \(A_i^U\).

\section{Conclusion}\label{sec:conclusion}

Privileged-likelihood scores should be treated as token credit only after checking outcome alignment, feedback construction, and how the loss uses them. On AIME~2025, our score was near random chance at separating correct from incorrect trajectories and all tested dense-reward configurations underperformed outcome-only GRPO, whereas stronger separation accompanied successful SDPO training on SciKnowEval Biology. Dense self-distillation can work, but the score must be validated before use.


\bibliography{references}
\bibliographystyle{iclr2027_conference}

\newpage
\appendix
\section{Supplementary Derivations and Protocol}\label{sec:appendix}

\subsection{Standing assumptions and scope}\label{app:assumptions}

Unless a result states otherwise, the following assumptions apply.
\begin{enumerate}
    \item \textbf{Actions and support.}  The local vocabulary \(\mathcal V\) is finite.
          Each result states the support it needs: forward KL requires
          \(q_{\mathrm{score}}(\cdot\mid s,c)\ll\pi_\theta(\cdot\mid s)\);
          likelihood ratios require positive common support
          for their numerator and denominator; reverse-KL centroid and Gibbs
          statements use full support where the optimized policy may place
          mass.  Termination is an action and is included when a token
          identity is lifted to a trajectory identity.
    \item \textbf{Utility.}  \(U(y)\) is integrable.  The action value
          \(Q^\pi(s,v)\) intervenes on \(v_t=v\) and then follows the
          declared continuation policy \(\pi\); it is not a teacher
          likelihood or an observational conditional unless the additional
          causal assumptions in \cref{app:outcome-conditionals} hold.
    \item \textbf{Local objectives.}  Frozen-target calculations hold the
          state/context weighting law, target \(\pi^\star\), token score \(d\) or \(d_\pm\),
          and reference policy \(\pi_{\mathrm{ref}}\) fixed.  They do not differentiate
          through target construction or state occupancy.
    \item \textbf{Geometry.}  Boundary rates use independent Euclidean
          softmax logits at one state.  They are not invariant to a neural
          parameterization, natural-gradient metric, optimizer
          preconditioning, or parameter sharing across states.
    \item \textbf{Detached estimators.}  Policy-gradient advantages,
          baselines, token weights, feedback, and scorer outputs are
          detached unless an upstream score term is explicitly shown.
          Expectations and derivatives may be interchanged.
    \item \textbf{Cross-fitting.}  Fold metadata is assigned by a
          content-independent random permutation after rollout and feedback
          generation but before donor selection and target scoring.  Donor
          and target randomness are conditionally independent, and the
          feedback judge's random seed is external to target generation.
\end{enumerate}

\subsection{Affine equivalence and coherent outcome conditionals}

\subsubsection{Proof of exact local affine equivalence}\label{app:affine-proof}

\paragraph{Proof.}
Suppress \(s\) in \(d\) and \(Q^\pi\). We use the full zero-sum
affine tangent space, corresponding to unrestricted local perturbations at an
interior policy.  Positive-affine equivalence says that, for one
\(\omega_1>0\) shared by all such perturbations,
\[
\langle \eta,d-\omega_1Q^\pi\rangle=0
\quad\text{for every }\eta\text{ such that }\mathbf{1}^\top \eta=0.
\]
The orthogonal complement of the zero-sum subspace in
\(\mathbb{R}^{|\mathcal V|}\) is
\(\operatorname{span}\{\mathbf{1}\}\).  Hence
\(d-\omega_1Q^\pi=\omega_0\mathbf{1}\), which is
\eqref{eq:affine}.  Conversely, the constant term vanishes against every
zero-sum perturbation, so \eqref{eq:affine} gives
\(\langle \eta,d\rangle=\omega_1\langle \eta,Q^\pi\rangle\).
\hfill\(\square\)

\paragraph{Ranking is insufficient.}
Let three actions have \(Q^\pi=(0,1,2)\) and \(d=(0,1,4)\). The rankings are
identical, but no positive affine map sends \(Q^\pi\) to \(d\). Indeed, for
\(\eta=(1,-2,1)\), \(\langle \eta,Q^\pi\rangle=0\) while
\(\langle \eta,d\rangle=2\).  AUC, rank correlation, or any strictly
increasing nonlinear transformation can therefore pass while exact local
functional equivalence fails.

\subsubsection{Outcome-conditional likelihood ratios}\label{app:outcome-conditionals}

\begin{prop}[Outcome-conditional odds identity; standard]
\label{prop:outcome-odds}
Suppose \(U\in\{0,1\}\), both outcome classes have positive probability at
state \(s\), and
\(q_u(v\mid s)=P(v_t=v\mid s_t=s,U=u)\) have common support.  Then
\begin{equation}\label{eq:outcome-odds-app}
\log\frac{q_1(v\mid s)}{q_0(v\mid s)}
=\operatorname{logit}P(U=1\mid s,v)
 -\operatorname{logit}P(U=1\mid s).
\end{equation}
If the action at \(s\) is randomized according to the declared policy,
consistency holds, there is no unmeasured action--outcome confounding
conditional on \(s\), positivity holds, and future actions follow \(\pi\),
then \(P(U=1\mid s,v)=Q^\pi(s,v)\).
\end{prop}

\paragraph{Proof.}
Bayes' rule yields
\[
\frac{P(v\mid s,U=1)}{P(v\mid s,U=0)}
=
\frac{P(U=1\mid s,v)}{P(U=0\mid s,v)}
\frac{P(U=0\mid s)}{P(U=1\mid s)}.
\]
Taking logs proves \eqref{eq:outcome-odds-app}.  The causal substitution is
the standard sequential identification step under the stated assumptions.
\hfill\(\square\)

This identity connects privileged likelihood to principled hindsight
credit assignment \citep{hca,cocoa}, but also locates the missing
assumption in free-form critique scoring.  A text labeled ``correct'' need
not be drawn from \(P(v\mid s,U=1)\), and two separately prompted texts
need not be conditionals of any common joint distribution.  Even when they
are coherent, the right side is a logit transform of binary outcome
probability.  It therefore gives ranking but does not generally guarantee the
affine equality in \cref{prop:affine}; special finite-action configurations,
including any strictly ordered two-action pair, can satisfy that equality
accidentally.

\subsection{KL identities}

\subsubsection{Soft-optimal trajectory ratios are a special case}

\begin{prop}[KL-regularized log-ratio telescoping; standard]
\label{prop:soft-telescope}
Let \(\pi_{\mathrm{ref}}(y\mid x)\) have full support, let autoregressive transitions be
deterministic given the token history, and let \(R(y,c)\) be a terminal
trajectory reward distinct from task utility \(U(y)\), with finite partition function
\(Z(x,c)=\mathbb E_{y\sim\pi_{\mathrm{ref}}(\cdot\mid x)}\exp(R(y,c)/\beta)\). Assume also that \(\mathbb E_{\pi^\star}|R(y,c)|<\infty\) for the Gibbs distribution defined below. The unrestricted optimum over a candidate distribution \(\pi\) of
\[
\max_\pi\;
\mathbb{E}_{y\sim\pi(\cdot\mid x,c)}R(y,c)
-\beta D_{\mathrm{KL}}\!\left(
\pi(\cdot\mid x,c)\,\|\,\pi_{\mathrm{ref}}(\cdot\mid x)\right),
\qquad \beta>0,
\]
is
\[
\pi^\star(y\mid x,c)
=\frac{\pi_{\mathrm{ref}}(y\mid x)\exp(R(y,c)/\beta)}{Z(x,c)}.
\]
For the complete autoregressive trajectory, including termination,
\begin{equation}\label{eq:soft-telescope}
\sum_t\log
\frac{\pi^\star(v_t\mid s_t,c)}{\pi_{\mathrm{ref}}(v_t\mid s_t)}
=\frac{R(y,c)}{\beta}-\log Z(x,c).
\end{equation}
Under the corresponding soft Bellman equations, each summand equals the
regularized soft advantage
\((Q_R^\star(s_t,v_t;c)-V_R^\star(s_t;c))/\beta\), where \(Q_R^\star\)
is the soft-optimal continuation value for \(R\) before the current action's
KL cost and
\(V_R^\star(s;c)=\beta\log\sum_u\pi_{\mathrm{ref}}(u\mid s)
\exp(Q_R^\star(s,u;c)/\beta)\). These values are defined by \(R\), not by
the task utility \(U\).
\end{prop}

\paragraph{Proof.}
The exponential-tilt optimizer follows from the Gibbs variational identity.
Taking the log ratio of its trajectory density to \(\pi_{\mathrm{ref}}\) and applying
the autoregressive chain rule gives \eqref{eq:soft-telescope}.  The
per-token identity follows by substituting the regularized Bellman
optimality equations for \(Q_R^\star\) and \(V_R^\star\).  These are standard KL-control and preference-
optimization identities \citep{regularizedmdp,dpo,donskervaradhan}.
\hfill\(\square\)

For fixed \((x,c)\), the term \(-\log Z(x,c)\) is constant across
trajectories.  A grouped prompt baseline removes it only when compared
trajectories share \(c\), or at least share the same normalizer, and when
sequence weighting, normalization, and clipping are compatible.  With a
different rollout-specific context \(c_i\) for each rollout, the normalizer can vary across
examples.
An arbitrary text-conditioned scorer need not be a soft-optimal policy and
need not satisfy \eqref{eq:soft-telescope}.  A terminal verdict also does not
uniquely determine how reward should be divided among tokens; a learned model
can infer such a division only by adding structure from data or prior
knowledge.

\subsubsection{Forward- and reverse-KL projections}
\label{app:projection-proof}

\begin{prop}[Forward-KL projection]\label{prop:fkl-projection}
For a fixed state \(s\) and context distribution, define the mean target
\(\bar q_{\mathrm{score}}(v\mid s)=\mathbb E_{c\mid s}q_{\mathrm{score}}(v\mid s,c)\). Here \(I(v;c\mid s)\) is computed under the joint distribution \(p(c\mid s)q_{\mathrm{score}}(v\mid s,c)\). Then
\begin{equation}\label{eq:fkl-projection}
\mathbb E_{c\mid s}D_{\mathrm{KL}}\!\left(q_{\mathrm{score}}(\cdot\mid s,c)\,\|\,\pi_\theta(\cdot\mid s)\right)
=I(v;c\mid s)+D_{\mathrm{KL}}\!\left(\bar q_{\mathrm{score}}(\cdot\mid s)\,\|\,\pi_\theta(\cdot\mid s)\right).
\end{equation}
The minimizing policy is \(\pi_\theta=\bar q_{\mathrm{score}}\), and the minimum is \(I(v;c\mid s)\).
\end{prop}

\paragraph{Proof.}
For a fixed state, add and subtract \(\log\bar q_{\mathrm{score}}(v\mid s)\):
\begin{align}
&\mathbb{E}_{c\mid s}
D_{\mathrm{KL}}\!\left(q_{\mathrm{score}}(\cdot\mid s,c)\,\|\,\pi_\theta(\cdot\mid s)\right) \\
&\quad =
\mathbb{E}_{c,v\mid s}
\log\frac{q_{\mathrm{score}}(v\mid s,c)}{\bar q_{\mathrm{score}}(v\mid s)}
+
\mathbb{E}_{c,v\mid s}
\log\frac{\bar q_{\mathrm{score}}(v\mid s)}{\pi_\theta(v\mid s)} \\
&\quad =
I(v;c\mid s)+D_{\mathrm{KL}}(\bar q_{\mathrm{score}}\|\pi_\theta),
\end{align}
where the second equality marginalizes \(c\) in the second term. Nonnegativity
of KL gives the minimizer and minimum. At a fixed state, forward KL differs
from cross entropy by a target-only constant, so differentiation through
independent softmax logits gives
\(\nabla_{\mathbf z}\mathbb E_{c\mid s}D_{\mathrm{KL}}(q_{\mathrm{score}}\|\pi_\theta)
=\pi_\theta(\cdot\mid s)-\bar q_{\mathrm{score}}(\cdot\mid s)\). In the full expectation this term is weighted
by the frozen state law.
\hfill\(\square\)

\paragraph{Reverse-KL centroid.}
Under common support and
\(\mathbb E_{c\mid s}|\log q_{\mathrm{score}}(v\mid s,c)|<\infty\)
for every \(v\),
\[
\begin{aligned}
&\mathbb{E}_{c\mid s}D_{\mathrm{KL}}\!\left(\pi_\theta(\cdot\mid s)\,\|\,q_{\mathrm{score}}(\cdot\mid s,c)\right)\\
&\quad=\sum_v\pi_\theta(v\mid s)\log\pi_\theta(v\mid s)\\
&\qquad-\sum_v\pi_\theta(v\mid s)\mathbb{E}_{c\mid s}\log q_{\mathrm{score}}(v\mid s,c).
\end{aligned}
\]
A Lagrange multiplier for \(\sum_v\pi_\theta(v\mid s)=1\) gives
\[
\pi_{\mathrm{geo}}(v\mid s)
=
\frac{\exp(\mathbb{E}_{c\mid s}\log q_{\mathrm{score}}(v\mid s,c))}
     {\sum_u\exp(\mathbb{E}_{c\mid s}\log q_{\mathrm{score}}(u\mid s,c))}.
\]
Thus the arithmetic-centroid/conditional-mutual-information identity is
specific to forward KL; it cannot be transferred unchanged to reverse KL.

\paragraph{Positive conditional mutual information without utility harm.}
Consider a one-step bandit. Let \(c\) be an independent style bit, and let
\(q_{\mathrm{score}}(\cdot\mid s,c)\) choose between two different but task-equivalent
surface-form actions. If terminal utility is one for both, then \(I(v;c\mid s)>0\), while
every target-supported action has the same task value.  The blind student's irreducible
forward-KL loss is positive, but the projection creates no utility regret.
Conversely, \(I(v;c\mid s)=0\) can occur when all contexts induce the same
uninformative target.  Either direction blocks a utility conclusion from
the projection scalar alone.

\subsection{Cross-fitting and partial policy gradients}\label{app:crossfit-proof}

\paragraph{Proof of \cref{prop:crossfit}.}
For the target-side calculation, freeze the donor law. By design,
\[
p(y^{\mathrm{don}},y^{\mathrm{tar}}\mid x,f)
=p(y^{\mathrm{don}}\mid x,f)\pi_{\mathrm{old}}(y^{\mathrm{tar}}\mid x,f).
\]
For external randomness \(\xi\) independent of \(y^{\mathrm{tar}}\) conditional on
\((x,f,y^{\mathrm{don}})\), and \(c=\phi(x,f,y^{\mathrm{don}},\xi)\),
marginalizing \((y^{\mathrm{don}},\xi)\) gives
\[
p(c,y^{\mathrm{tar}}\mid x,f)
=p(c\mid x,f)\pi_{\mathrm{old}}(y^{\mathrm{tar}}\mid x,f),
\]
so \(c\perp y^{\mathrm{tar}}\mid x,f\). Conditional on frozen \(c\), assume
each \(r_{c,k}\) is determined by the target history through token \(k\) and
does not depend on later target tokens. Standard score-function
differentiation and zero conditional covariance with earlier rewards then
give the target-side gradient below, where
\(R_c(y)=\sum_k r_{c,k}\) is the feedback-defined return:
\begin{equation}\label{eq:crossfit-gradient}
\nabla_\theta\mathbb E_{y^{\mathrm{tar}}\sim\pi_\theta}
\!\left[R_c(y^{\mathrm{tar}})\mid c\right]
=\mathbb E\!\left[
\sum_tG_{c,t}\nabla_\theta\log\pi_\theta(v_t\mid s_t)\mid c\right],
\qquad G_{c,t}=\sum_{k\geq t}r_{c,k},
\end{equation}
assuming no additional direct \(\theta\)-dependence \citep{reinforce,scg}. If a larger objective also
changes how donors or feedback are generated, this expression is only the
part of its gradient that passes through the target rollout.
\hfill\(\square\)

\paragraph{Upstream dependence.}
If the donor is policy generated,
\(y^{\mathrm{don}}\sim\pi_\theta(\cdot\mid x,f)\), and
\(c=\phi(y^{\mathrm{don}},\xi)\), then the intended joint
objective has an upstream donor score term
\[
\mathbb{E}_{y^{\mathrm{don}},y^{\mathrm{tar}},\xi}\!\left[
R_{\phi(y^{\mathrm{don}},\xi)}(y^{\mathrm{tar}})
\nabla_\theta\log\pi_\theta(y^{\mathrm{don}}\mid x,f)\right],
\]
plus any allowed pathwise derivatives.  Cross-fitting removes a target's
direct ancestry from its feedback; it does not erase policy dependence in
donor generation.

\paragraph{What own-rollout agreement alone cannot show.}
Let a sampled action \(v\in\{0,1\}\) write its own feedback \(c=v\), and
score it by \(R_c(v)=\mathbf{1}\{v=c\}\). Every realized action receives
one.  The feedback perfectly matches the sample yet defines no preference
between actions and can be independent of task utility.  Thus, own-rollout
agreement alone cannot validate transferred action value.  Held-out-feedback
construction removes this direct dependence and asks a different question.

\paragraph{Independent but task-harming feedback.}
Let an independent donor \(y^{\mathrm{don}}\sim\operatorname{Bernoulli}(0.9)\)
set \(c=y^{\mathrm{don}}\), and reward a target action \(v\) by
\(\mathbf{1}\{v=c\}\). Then \(c\perp v\) before the target is sampled, but
the held-out-feedback objective favors \(v=1\). If true utility is
\(\mathbf{1}\{v=0\}\), optimizing that objective reduces task reward.
Independence therefore tells us which objective is being optimized, not
whether it is the right one.

\paragraph{Instantaneous versus future credit.}
Replacing \(G_{c,t}\) by the instantaneous \(r_{c,t}\) is exact only when
\[
\mathbb{E}\!\left[
\nabla_\theta\log\pi_\theta(v_t\mid s_t)
\sum_{k>t}r_{c,k}
\right]=0.
\]
Conditional action-independence of expected future reward is sufficient,
but not necessary.  A dense score at each position does not by itself
justify omitting downstream effects.

\subsection{Token-logit gradients for frozen targets}\label{app:geometry-proof}

\paragraph{General chain rule.}
Suppressing the fixed state \(s\), for token-logit vector \(\mathbf z\), the softmax Jacobian is
\[
\frac{\partial\pi_\theta(v')}{\partial z_v}
=\pi_\theta(v')(\mathbf{1}\{v'=v\}-\pi_\theta(v)).
\]
Therefore
\begin{equation}\label{eq:softmax-chain}
\frac{\partial\mathcal L}{\partial z_v}
=\sum_{v'}\frac{\partial\mathcal L}{\partial\pi_\theta(v')}
       \frac{\partial\pi_\theta(v')}{\partial z_v}
=\pi_\theta(v)\left(
\frac{\partial\mathcal L}{\partial\pi_\theta(v)}
-\sum_{v'}\pi_\theta(v')\frac{\partial\mathcal L}{\partial\pi_\theta(v')}\right),
\end{equation}

\paragraph{Reverse KL.}
For \(\mathcal L=D_{\mathrm{KL}}(\pi_\theta\|\pi^\star)\),
\(\partial\mathcal L/\partial\pi_\theta(v)=\log(\pi_\theta(v)/\pi^\star(v))+1\), hence
\[
\frac{\partial\mathcal L}{\partial z_v}
=\pi_\theta(v)\left[\log\frac{\pi_\theta(v)}{\pi^\star(v)}
-D_{\mathrm{KL}}(\pi_\theta\|\pi^\star)\right].
\]
With two actions, let \(p=\pi_\theta(v\mid s)=1-\varepsilon\) and denote the other
token by \(\bar v\). Direct cancellation gives
\[
\frac{\partial\mathcal L}{\partial z_v}
=p(1-p)\log\frac{p \pi^\star(\bar v)}{(1-p)\pi^\star(v)}.
\]
For fixed positive \(\pi^\star(v),\pi^\star(\bar v)\), this is
\(\Theta(\varepsilon\log(1/\varepsilon))\).

\paragraph{Generalized JSD.}
For \(0<\lambda<1\),
\[
\mathcal L=\lambda D_{\mathrm{KL}}(\pi_\theta\|m)
 +(1-\lambda)D_{\mathrm{KL}}(\pi^\star\|m),
\qquad m=\lambda\pi_\theta+(1-\lambda)\pi^\star,
\]
differentiation, including the dependence of \(m\) on \(\pi_\theta\), simplifies
to
\(\partial\mathcal L/\partial\pi_\theta(v)=\lambda\log(\pi_\theta(v)/m(v))\). In the two-action
case,
\[
\frac{\partial\mathcal L}{\partial z_v}
=\lambda p(1-p)
\log\frac{p m(\bar v)}{(1-p)m(v)}.
\]
Because \(m(\bar v)\to(1-\lambda)\pi^\star(\bar v)>0\), this has the same
\(\Theta(\varepsilon\log(1/\varepsilon))\) order.

\paragraph{Bounded reward.}
For loss \(\mathcal L=-\sum_{v'}\pi_\theta(v')d_\pm(v')\),
\[
\frac{\partial\mathcal L}{\partial z_v}
=p(1-p)\left(d_\pm(\bar v)-d_\pm(v)\right).
\]
It is \(O(\varepsilon)\) for bounded \(d_\pm\) and
\(\Theta(\varepsilon)\) when the token-score gap
\(d_\pm(\bar v)-d_\pm(v)\) is bounded away from zero.  The direction is corrective
only when the score favors the desired alternative.  A state-only baseline
changes the realized advantage but not this expected gradient.

\paragraph{Forward KL.}
For \(\mathcal L=D_{\mathrm{KL}}(\pi^\star\|\pi_\theta)\),
\(\partial\mathcal L/\partial\pi_\theta(v)=-\pi^\star(v)/\pi_\theta(v)\). Substitution into
\eqref{eq:softmax-chain} yields
\(\partial\mathcal L/\partial z_v=\pi_\theta(v)-\pi^\star(v)\), and therefore
\(\partial\mathcal L/\partial z_v\to1-\pi^\star(v)\).

\paragraph{Scope of the singularity comparison.}
If near the relevant boundary
\(\mathcal L(\pi_\theta)=-\sum_v \rho(v)\log\pi_\theta(v)+O(1)\), with fixed weights and a
remainder whose logit gradient vanishes, then \(\mathcal L\) has forward-KL-like
leading logit behavior.  Faster singularities can also cancel or dominate the
softmax Jacobian.  The four losses derived above are
compared in this common Euclidean-logit coordinate system.

\begin{figure}[t]
\centering
\includegraphics[width=\linewidth]{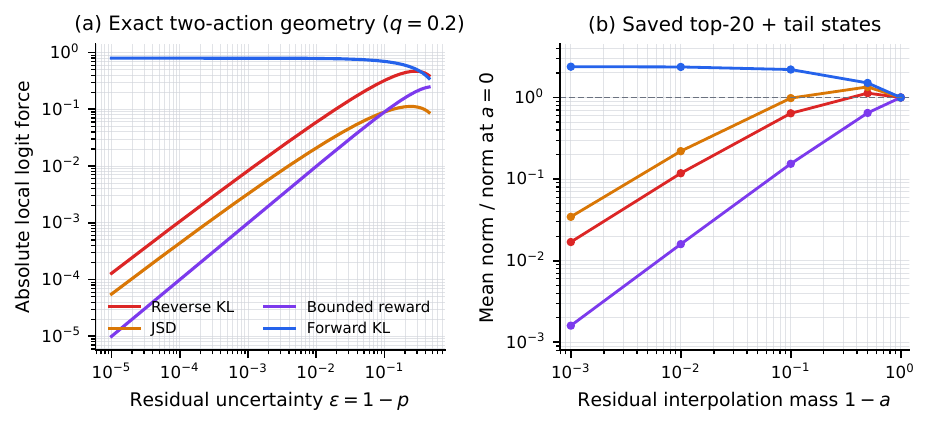}
\caption{\textbf{Token-logit gradients for a frozen target.}
The left panel evaluates exact two-action Euclidean-logit gradients as
student confidence approaches one while a fixed target retains off-token
mass.  The right panel repeats the interpolation on \(74{,}653\) saved
top-20-plus-tail distributions from the three dense-score configurations,
with the remaining probability mass aggregated into one tail bin. Here \(a\in[0,1]\) is the interpolation
weight toward the most likely token; the left panel's \(q=0.2\) denotes
\(\pi^\star(\bar v\mid s)=0.2\). At \(a=0.999\), state-count-weighted
Euclidean-logit gradient-norm ratios, normalized by their values at \(a=0\),
are \(0.0169\), \(0.0345\), \(0.0016\), and \(2.384\) for reverse KL,
generalized JSD, bounded reward, and forward KL. Constant loss rescaling
changes their absolute magnitudes.}
\label{fig:gradient-geometry}
\end{figure}

\subsection{Scalar token weights and expected parameter gradients}
\label{app:update-counterexample}

\paragraph{Proof of \cref{prop:gradient}.}
By definition,
\begin{equation}\label{eq:uniform-gradient}
g_U=\mathbb E\!\left[
\frac{A_i^U}{T_i}\sum_t h_{i,t}\right]
\end{equation}
and
\[
g_A
=\mathbb{E}\!\left[
\frac{A_i^U}{T_i}\sum_t w_{i,t}h_{i,t}\right].
\]
Therefore
\begin{equation}\label{eq:gradient-difference}
g_A-g_U=\mathbb E\!\left[
\frac{A_i^U}{T_i}\sum_t(w_{i,t}-1)h_{i,t}\right].
\end{equation}
Mean-one normalization only states
\(\sum_t(w_{i,t}-1)=0\); it does not state that the weighted sum of the
generally distinct vectors \(h_{i,t}\) is zero. \hfill\(\square\)

\paragraph{Two-coordinate counterexample.}
Let \(v_1,v_2\) be independent Bernoulli actions with separate logits and
\(P(v_1=1)=P(v_2=1)=1/2\). Set
\[
A^U=v_1+v_2-1,\qquad
(w_1,w_2)=(2-\varepsilon,\varepsilon),\quad 0<\varepsilon<1.
\]
The weights are positive and have mean one.  Since the score for coordinate
\(t\) is \(v_t-1/2\),
\[
g_U=(1/8,1/8),\qquad
g_A=((2-\varepsilon)/8,\varepsilon/8).
\]
The trajectory advantage keeps its sign and
\((A^U/2)(w_1+w_2)=A^U\), yet the expected policy-gradient vector
changes.

\paragraph{Invariants that do survive.}
Before later clipping or normalization, \(w_{i,t}>0\) preserves each
token advantage's sign, \(A_i^U=0\) remains zero, and mean-one weights
preserve the scalar sequence budget.  At a fixed, commonly observed position
\(t\), they do not preserve a prompt-group mean because, even if
\(\mathbb{E}_{i\mid x}A_i^U=0\),
\[
\mathbb{E}_{i\mid x}[A_i^Uw_{i,t}]
=\operatorname{Cov}_{i\mid x}(A_i^U,w_{i,t})
\]
need not vanish.  Variable-length masks must be included when the position is
not common to all trajectories.  The weights also do not preserve the
expected policy-gradient estimate.  If trajectory- or policy-dependent weights
are recomputed and then detached,
the resulting update rule need not be the gradient of any scalar
objective.  Thus, preserving scalar sign and total weight does not preserve
the parameter update in verifier-anchored self-distillation
\citep{rlsd,rlcsd}.

\paragraph{Relation to reward redistribution.}
Valid return redistribution preserves a specified return or optimal-policy
relation under explicit conditions \citep{rudder}.  Arbitrarily inserting
mean-normalized weights into a score-function sum is not equivalent to
constructing a reward process whose return-to-go yields the same policy
gradient. The score vectors, not only the scalar total, determine that
equivalence.

\subsection{Sequence length and local optima}

\begin{prop}[Elementary variable-length aggregation lemma]
\label{prop:variable-horizon}
For each sequence length \(T\), let
\[
\operatorname{Agg}_T(A^C)=\sum_{t=1}^T w_{T,t}A^C_t,\qquad w_{T,t}\geq0.
\]
Here \(\mathbf{1}\) is the all-ones vector and \(e_j\) is the \(j\)-th standard basis vector. Fix \(\mu\neq0\) and \(\delta>0\). There are no constants \(M<\infty\)
and \(\nu>0\), uniform in \(T\), such that both
\[
\left|\operatorname{Agg}_T(\mu\mathbf{1})\right|\leq M
\quad\text{and}\quad
\operatorname{Agg}_T(\delta e_j)\geq \nu\delta
\quad\text{for every }j\leq T\text{ and every }T.
\]
\end{prop}

\paragraph{Proof.}
The singleton condition gives \(w_{T,j}\geq \nu\) for every \(j\). Hence
\(\sum_t w_{T,t}\geq \nu T\), and
\[
\left|\operatorname{Agg}_T(\mu\mathbf{1})\right|
=|\mu|\sum_t w_{T,t}\geq|\mu|\nu T,
\]
contradicting a uniform bound. \hfill\(\square\)

For the nonnegative linear aggregators defined in
\cref{prop:variable-horizon}, a sum can retain an isolated event but
accumulates a repeated offset as the sequence grows; a mean bounds that
offset but gives an \(O(1/T)\) response to a fixed number of events.

\begin{prop}[A fixed detached token score has a boundary optimum]
\label{prop:fixed-field}
At a fixed state, the maximizers of
\[
\sum_v\pi_\theta(v\mid s)d_\pm(s,v)
\]
over the probability simplex are exactly the distributions supported on
\(\arg\max_v d_\pm(s,v)\).  With a unique maximizing token, the optimizer is a
point mass.
\end{prop}

\paragraph{Proof.}
The displayed objective is a linear functional on the simplex. Any mass on an action below
the maximum can be moved to a maximizing action to increase the objective.
\hfill\(\square\)

A strictly increasing, policy-independent pointwise transformation preserves
token ordering and the same boundary maximizers; non-strict clipping can
introduce ties.

\subsection{KL regularization creates a finite target, not a correct one}
\label{app:gibbs-tether}

\begin{prop}[KL-regularized optimum; standard]\label{prop:gibbs-tether}
Fix a finite two-sided score \(d_\pm(s,\cdot)\), a full-support reference policy
\(\pi_{\mathrm{ref}}(\cdot\mid s)\),
\(\alpha\in\mathbb{R}\), and \(\beta>0\).  Define
\[
F_{\mathrm{KL}}(\pi_\theta;s)
=\alpha\mathbb{E}_{v\sim\pi_\theta(\cdot\mid s)}d_\pm(s,v)
-\beta D_{\mathrm{KL}}\!\left(\pi_\theta(\cdot\mid s)\,\|\,\pi_{\mathrm{ref}}(\cdot\mid s)\right).
\]
Then
\begin{align}
\pi^\star(v\mid s)
 &=
\frac{\pi_{\mathrm{ref}}(v\mid s)\exp((\alpha/\beta)d_\pm(s,v))}
     {Z_s}, \\
F_{\mathrm{KL}}(\pi_\theta;s)
 &=-\beta D_{\mathrm{KL}}\!\left(\pi_\theta(\cdot\mid s)\,\|\,\pi^\star(\cdot\mid s)\right)
   +\beta\log Z_s ,
\end{align}
where
\(Z_s=\sum_v\pi_{\mathrm{ref}}(v\mid s)\exp((\alpha/\beta)d_\pm(s,v))\).
Thus \(\pi^\star\) is the unique maximizer.
\end{prop}

\paragraph{Proof.}
Substitute
\[
\log\pi^\star(v\mid s)
=\log \pi_{\mathrm{ref}}(v\mid s)+(\alpha/\beta)d_\pm(s,v)-\log Z_s
\]
into \(D_{\mathrm{KL}}(\pi_\theta\|\pi^\star)\) and rearrange.  Uniqueness follows
from strict convexity of KL in its first argument.  This is the standard
Gibbs/Donsker--Varadhan variational identity used in KL-regularized control
\citep{donskervaradhan,regularizedmdp}. \hfill\(\square\)

The result guarantees a finite local target, not a good one. Its target is
defined jointly by \(\pi_{\mathrm{ref}}\) and the uncalibrated score \(d_\pm\);
it can be a finite-confidence target for the wrong action. If either object is
refreshed, \(\pi^\star\) is a per-update target rather than a demonstrated
global equilibrium.

For the implemented KL configurations, let
\[
d_\pm^{\mathrm{clip}}(s,v)=\operatorname{clip}(d_\pm(s,v),-2,2),
\qquad
\ell(s,v)=\log\frac{\pi_{\mathrm{old}}(v\mid s)}{\pi_{\mathrm{ref}}(v\mid s)}.
\]
Their clipped and full token advantages are
\begin{align}
A^{C,\mathrm{clip}}_{i,t}
&=\operatorname{ctr}_{s_{i,t}}[d_\pm^{\mathrm{clip}}](v_{i,t}),\\
A^{C,\mathrm{full}}_{i,t}
&=\operatorname{ctr}_{s_{i,t}}
[\alpha d_\pm^{\mathrm{clip}}-\beta\ell](v_{i,t}).
\label{eq:full-kl-rule}
\end{align}
For the projected-KL variant, let
\begin{equation}\label{eq:projected-kl-slope}
\widehat\gamma_s=
\frac{\operatorname{Cov}_{v\sim\pi_{\mathrm{old}}(\cdot\mid s)}
\!\left(d_\pm^{\mathrm{clip}}(s,v),\ell(s,v)\right)}
{\operatorname{Var}_{v\sim\pi_{\mathrm{old}}(\cdot\mid s)}
\!\left(d_\pm^{\mathrm{clip}}(s,v)\right)}.
\end{equation}
Set \(\widehat\gamma_s=0\) when the denominator is zero, and let the configured scale be
\(\kappa>0\). Its token advantage is
\begin{equation}\label{eq:projected-kl-rule}
A^{C,\mathrm{proj}}_{i,t}
=\kappa(\alpha-\beta\widehat\gamma_{s_{i,t}})
A^{C,\mathrm{clip}}_{i,t}.
\end{equation}
The projected rule in \eqref{eq:projected-kl-rule}
retains only the least-squares component of the reference-policy correction
that varies with the centered feedback score.  Because it keeps that
score-aligned correction and discards the orthogonal remainder, it need not
correspond to this full-distribution objective.

\subsection{Baseline details}

At the on-policy point \(\pi_\theta=\pi_{\mathrm{old}}\), let
\(d_\pm(s,\cdot)\) be fixed before \(v\sim\pi_{\mathrm{old}}(\cdot\mid s)\)
is sampled, and let \(h(v)=\nabla_\theta\log\pi_\theta(v\mid s)\). Since
\(\mathbb{E}_{v\sim\pi_{\mathrm{old}}}[h(v)\mid s]=0\), every state-only scalar baseline \(b(s)\)
satisfies
\[
\mathbb{E}[(d_\pm(s,v)-b(s))h(v)\mid s]
=\mathbb{E}[d_\pm(s,v)h(v)\mid s].
\]
This cancellation is the unclipped on-policy score-function identity for a
score fixed before sampling. If the context \(c(y)\) is written from the
sampled rollout \(y\), the score is trajectory dependent, so the identity
applies only after conditioning on a fixed feedback context.
The baseline minimizing the scalar residual
\(\mathbb{E}[(d_\pm(s,v)-b(s))^2\mid s]\) is
\(b(s)=\mathbb{E}[d_\pm(s,v)\mid s]\), which is
\eqref{eq:same-slot-baseline}.  The baseline minimizing the trace of the
one-state vector estimator's covariance is instead
\[
b^\star(s)
=\frac{\mathbb{E}[d_\pm(s,v)\|h(v)\|^2\mid s]}
       {\mathbb{E}[\|h(v)\|^2\mid s]}.
\]
These coincide only under additional conditions. For sampled token \(v\)
with \(p=\pi_{\mathrm{old}}(v\mid s)\), the same-slot baseline also exposes
\[
d_\pm(s,v)-\mathbb{E}_{u\sim\pi_{\mathrm{old}}(\cdot\mid s)}d_\pm(s,u)
=(1-p)\left(d_\pm(s,v)
-\frac{\sum_{u\neq v}\pi_{\mathrm{old}}(u\mid s)d_\pm(s,u)}{1-p}\right).
\]
This factorization describes a realized centered advantage. The
vanishing expected softmax-logit gradients in
\cref{prop:boundary-geometry} do not depend on choosing this particular
baseline.

\subsection{Additional protocol specification}\label{app:protocol}

\subsubsection{One-sided score audit}\label{app:one-sided-audit}

For a fixed normalized scorer \(q_{\mathrm{score}}(\cdot\mid c^+)\), rollout
policy \(\pi_{\mathrm{old}}\), and the set \(\mathcal G\) of correct trajectories,
\[
\mathbb E_{\pi_{\mathrm{old}}(\cdot\mid \mathcal G)}
\log\frac{q_{\mathrm{score}}(y\mid c^+)}{\pi_{\mathrm{old}}(y)}
=
\log\frac{q_{\mathrm{score}}(\mathcal G\mid c^+)}
{\pi_{\mathrm{old}}(\mathcal G)}
-D_{\mathrm{KL}}\!\left(
\pi_{\mathrm{old}}(\cdot\mid \mathcal G)
\,\|\,q_{\mathrm{score}}(\cdot\mid c^+,\mathcal G)\right).
\]
The outcome-conditioned mean therefore combines the change in teacher mass
on \(\mathcal G\) with the within-\(\mathcal G\) KL term. A negative
conditional mean establishes lower geometric likelihood on the observed
successful paths; the first term separately determines the teacher's total
mass on \(\mathcal G\). With rollout-specific feedback \(c^+=c^+(y)\), the diagonal
scores do not in general form one normalized teacher distribution.

\Cref{tab:one-sided-fixed-audit} reconstructs the positive teacher leg on
fixed rollouts from two configurations that use the two-sided score. In both
analyses, more than \(99\%\) of regraded correct trajectories have a negative
complete-trajectory teacher--policy log ratio.

\begin{table}[H]
\centering
\caption{\textbf{Fixed-rollout one-sided audit.}
\(\sum_t d(s_t,v_t)\) is the complete-trajectory one-sided teacher--policy
log ratio. The correct and incorrect columns report the percentage of each
class with a negative ratio. AUC uses the response-mean token residual, with
95\% prompt-bootstrap intervals. These rows extract the positive leg from
two experiments that use the two-sided score.}
\label{tab:one-sided-fixed-audit}
\small
\setlength{\tabcolsep}{3pt}
\begin{tabular}{@{}lccc@{}}
\toprule
Configuration & Correct & Incorrect & AUC [95\% CI] \\
\midrule
Source-clipped additive trajectories
  & \(99.49\%\)
  & \(99.89\%\)
  & \(0.435\;[0.393,\,0.474]\) \\
Saliency-modulated trajectories
  & \(99.28\%\)
  & \(99.51\%\)
  & \(0.512\;[0.470,\,0.547]\) \\
\bottomrule
\end{tabular}
\end{table}

\subsubsection{Score provenance}\label{app:score-provenance}

\Cref{tab:score-provenance} defines every score in
\cref{tab:score-auc,tab:secondary-score-auc}.  The external LLM judge writes
feedback; \(q_{\mathrm{score}}\) is instantiated by the listed
\texttt{gpt-oss-20b} checkpoint and supplies all displayed privileged token
likelihoods.

\begin{table}[H]
\centering
\caption{\textbf{Score definitions and model provenance.}
``Frozen/earlier'' means that the scoring checkpoint is not updated through
the score being measured.}
\label{tab:score-provenance}
\scriptsize
\begin{tabularx}{\linewidth}{@{}L{0.21\linewidth}L{0.48\linewidth}X@{}}
\toprule
Displayed score & Feedback construction and writer & Likelihood scorer \\
\midrule
Own-rollout feedback
& Rollout-specific helpful and adversarial feedback; \texttt{gpt-5.6-luna}
& frozen/earlier \texttt{gpt-oss-20b} \\
Implemented additive
& Value-masked positive feedback against a same-problem mixture of plausible near-miss negatives; \texttt{gpt-5.6-luna}
& frozen reference \texttt{gpt-oss-20b} \\
Token-weighting configuration, score sum
& Value-masked positive feedback and plausible near-miss negative feedback; \texttt{gpt-5.6-luna}
& frozen/earlier \texttt{gpt-oss-20b} \\
Held-out masked feedback, sum
& Value-masked/near-miss feedback from opposite-fold rollouts of the same problem; \texttt{gpt-5.6-luna}
& frozen \texttt{gpt-oss-20b} \\
Held-out masked feedback, mean
& Same feedback as the preceding row, reduced by token mean; \texttt{gpt-5.6-luna}
& frozen \texttt{gpt-oss-20b} \\
Held-out correct-fact feedback, mean
& Correct versus incorrect interior facts from opposite-fold rollouts of the same problem; \texttt{gpt-5.6-luna}
& frozen \texttt{gpt-oss-20b} \\
Different-problem correct-fact control
& The preceding correct-fact construction with feedback from a different problem; \texttt{gpt-5.6-luna}
& frozen \texttt{gpt-oss-20b} \\
Identical-feedback null
& Identical positive and negative contexts; \texttt{gpt-5.6-luna}
& frozen \texttt{gpt-oss-20b} \\
\bottomrule
\end{tabularx}
\end{table}

\begin{table}[H]
\centering
\caption{\textbf{Secondary trajectory-score diagnostics.}
Entries are AUCs with 95\% prompt-bootstrap confidence intervals.}
\label{tab:secondary-score-auc}
\scriptsize
\setlength{\tabcolsep}{2.6pt}
\begin{tabular}{@{}lccc@{}}
\toprule
Score construction
  & Raw pooled AUC
  & Length-adjusted AUC
  & Within-prompt AUC \\
\midrule
Token-weighting configuration, score sum
  & \(0.553\;[0.517,\,0.585]\)
  & \(0.528\;[0.493,\,0.560]\)
  & \(0.547\;[0.506,\,0.583]\) \\
Held-out masked feedback, sum
  & \(0.453\;[0.384,\,0.526]\)
  & \(0.489\;[0.431,\,0.555]\)
  & \(0.520\;[0.463,\,0.584]\) \\
Different-problem correct-fact control
  & \(0.475\;[0.418,\,0.530]\)
  & \(0.528\;[0.467,\,0.583]\)
  & \(0.520\;[0.458,\,0.577]\) \\
\bottomrule
\end{tabular}
\end{table}

\paragraph{Matched responses and format sensitivity.}
The own-rollout analysis uses \(3{,}361\) responses with a unique trajectory
match.

\Cref{tab:boxed-complete-auc} reports a format/completion sensitivity analysis
among responses that both completed and contained a boxed answer. Because
completion and formatting are post-generation behavior, this condition
changes the population being ranked. Intervals use the same prompt-bootstrap
procedure as \cref{tab:score-auc}.

\begin{table}[H]
\centering
\caption{\textbf{Raw AUC among boxed, completed responses.}
Brackets are 95\% prompt-bootstrap confidence intervals.}
\label{tab:boxed-complete-auc}
\small
\begin{tabularx}{0.78\linewidth}{@{}Xr@{}}
\toprule
Score & AUC [95\% CI] \\
\midrule
Own-rollout feedback & \(0.595\;[0.510,\,0.675]\) \\
Implemented additive & \(0.506\;[0.486,\,0.528]\) \\
Token-weighting configuration, score sum & \(0.552\;[0.518,\,0.588]\) \\
Held-out masked feedback, sum & \(0.451\;[0.382,\,0.515]\) \\
Held-out masked feedback, mean & \(0.514\;[0.451,\,0.573]\) \\
Held-out correct-fact feedback, mean & \(0.416\;[0.344,\,0.490]\) \\
Different-problem correct-fact control & \(0.472\;[0.417,\,0.526]\) \\
Identical-feedback null & \(0.503\;[0.499,\,0.507]\) \\
\bottomrule
\end{tabularx}
\end{table}

The token-weighting row in \cref{tab:secondary-score-auc} reports a sum of
signed contrasts, whereas the training weights use the mean contrast
magnitude.

\paragraph{Label and feedback audits.}
Among 49 gradeable strong-consensus cases from a targeted audit, 18
\((36.7\%)\) were equivalent to the reference. This estimate is conditional
on the selected strong-consensus cases and is not a corpus-wide label-error
rate.

An external-judge pilot selected the better solution in \(94.4\%\)
\((51/54)\) of valid judgments.

The \(c^+\) and \(c^-\) prompts also differ slightly beyond their mathematical
content. The positive template masks selected values, asks the model to work
them out, and more often includes formatting instructions, whereas the
negative template supplies explicit near-miss values. The likelihood contrast
can therefore reflect template wording as well as the intended mathematical
distinction.

\subsubsection{Cross-fitted construction}

Within each prompt group, fold labels are assigned by a seeded,
content-independent permutation after the rollouts and their feedback records
have been generated, but before donor selection and target scoring.  Donor
feedback for a target fold comes only from trajectories in the other fold.  A
target rollout is excluded from every record used to write its feedback.  If
too few donors exist, the example is marked unavailable rather than silently
using feedback from the target itself.  Donor count and aggregation (one
donor, a fixed pool, or a declared log-mean-exp over negatives) are reported
with each score.

We verify that swapping \(c^+\) and \(c^-\) negates finite contrasts, setting
\(c^+=c^-\) produces a numerical null, and the scored rollout never appears
among the records used to write its feedback. Different-problem donors and
template-only pairs serve as controls for problem-specific information and
wording.

\subsubsection{Score reduction and training-rule configurations}\label{app:training-rule-configs}

Raw sequence sums, token means, centered or squashed scores, and own-rollout
or cross-fitted scores are distinct measurements.
\Cref{tab:training-rule-configs} specifies the feedback reduction, clipping,
gating, and combination with outcome credit for each training configuration.
Every matched configuration uses
\(A_i^U=U_i-\frac{1}{K-1}\sum_{j\ne i}U_j\), with the same length and validity
handling, and the outcome-only control sets the token advantage to zero.
The same-state baseline in \eqref{eq:same-slot-baseline} was approximated with
the scorer's top-\(k\) tokens plus one bin for all remaining probability mass.
The table states whether that aggregate tail-bin contribution was included.

\begin{table}[H]
\centering
\caption{\textbf{Training configurations.}
All configurations share the base training recipe, post-combination advantage
clipping, and subsequent policy-ratio clipping. Each row reports the evaluated
training recipe, including the jointly specified choices in the KL-regularized
variants.}
\label{tab:training-rule-configs}
\small
\begin{tabularx}{\linewidth}{@{}L{0.23\linewidth}X@{}}
\toprule
Configuration & How the token score enters training \\
\midrule
Outcome-only
& Uses \(A_i^U\); token advantage identically zero. \\
Entropy-gated additive
& Uses \(0.5\tanh(A^C_{i,t}/0.5)\), multiplies it by
  \(\min(1,H_{i,t}/0.2)\), gates it on nonzero \(A_i^U\), prevents it from
  reversing the sign of \(A_i^U\), and omits the aggregate tail bin. \\
Source-clipped additive
& Clips every \(d_\pm(s_{i,t},v)\) to \([-2,2]\) before centering, then uses
  \(0.5\tanh(A^C_{i,t}/0.5)\); gates it on nonzero \(A_i^U\), prevents reversal of
  the sign of \(A_i^U\), and omits the aggregate tail bin. \\
Entropy-gated + source-clipped
& Combines the preceding source clip, \(\tanh\) reduction, and entropy
  multiplier; gates on nonzero \(A_i^U\), prevents reversal of its sign, and
  omits the aggregate tail bin. \\
Full KL-regularized
& Uses \eqref{eq:full-kl-rule} with \(\alpha=\beta=1\), uses no \(\tanh\), gates on
  nonzero \(A_i^U\), and includes the aggregate tail bin. \\
Projected KL correction
& Uses \eqref{eq:projected-kl-rule} with
\(\alpha=\beta=1\) and \(\kappa=0.25\), uses no \(\tanh\), uses the same
  nonzero-\(A_i^U\) gate, and includes the aggregate tail bin. \\
\bottomrule
\end{tabularx}
\end{table}

The execution order is: construct \(A_i^U\) and \(A^C_{i,t}\); apply
the dense gate and any entropy multiplier; combine the channels; apply the
additive sign clamp where configured; clip the combined advantage to
\([-1,1]\); mask response tokens; and finally apply PPO importance-ratio
clipping in the actor loss.  Advantage normalization is disabled.
Final advantage clipping can discard an added score when the utility
advantage is already at the clip limit, or reduce it when the added score
has the opposite sign.

\paragraph{Mechanism analysis.}
In a separately analyzed ungated additive configuration, \(57\%\)--\(71\%\)
of absolute scalar weight fell in the highest-entropy token decile. Across
the matched configurations through update~25, the additive variants moved
toward higher token entropy and reference-policy KL, while the KL-regularized
variants remained near the outcome-only range.

\subsubsection{Prompt-level comparisons and uncertainty}

For training configuration \(\ell\), prompt \(x\), update \(n\), and rollout \(i\), let \(\widehat U_{\ell,x,n,i}\in\{0,1\}\) be the corrected equivalent-answer label and let
\[
\widehat p_{\ell,x,n}
=\frac{1}{K_{\ell,x,n}}\sum_{i=1}^{K_{\ell,x,n}}
\widehat U_{\ell,x,n,i}
\]
be the corresponding prompt accuracy. The primary paired difference
at a fixed update is
\[
\widehat\Delta_n
=\frac{1}{|\mathcal X_n|}
\sum_{x\in\mathcal X_n}
(\widehat p_{\ell,x,n}-\widehat p_{0,x,n}),
\]
where subscript \(0\) denotes the outcome-only control and \(\mathcal X_n\) contains prompts observed in both configurations. Prompt-cluster bootstrap samples prompts, carrying
all their rollouts together. The matched comparison uses one seed per
configuration, and its checkpoints are repeated measurements from that seed.

The paired configuration-label sign-flip test enumerates the prompt-level
differences under a sharp exchangeability null for the two configuration
labels within each paired prompt. The reported \(p\)-value is the exact tail
probability under that null.

Across the ten method-by-update contrasts, the Bonferroni threshold is
\(0.005\); all five update~49 exact \(p\)-values remain below this threshold.
The reported confidence intervals are unadjusted.

\paragraph{Completion check.}
At update~49, the outcome-only configuration was \(76.7\%\) boxed and complete
with \(23.3\%\) truncation, whereas the five token-score configurations were
\(94.2\%\) to \(98.3\%\) boxed, \(95.8\%\) to \(100\%\) complete, and
\(0\%\) to \(4.2\%\) truncated. The lower correctness of the token-score
configurations occurs despite their higher completion and lower truncation.

For outcome-ranking analyses, the bootstrap likewise carries all
trajectories from a prompt together.  We report pooled AUC, within-prompt
AUC, score--length association, a predeclared length/format/completion
adjustment, and parameter-gradient projection as separate statistics because
they answer different questions.
AUC screens outcome ranking, while \cref{prop:affine} characterizes exact
local equivalence.

\subsubsection{Shared-checkpoint selection}

\begin{figure}[!t]
\centering
\includegraphics[width=0.94\linewidth]{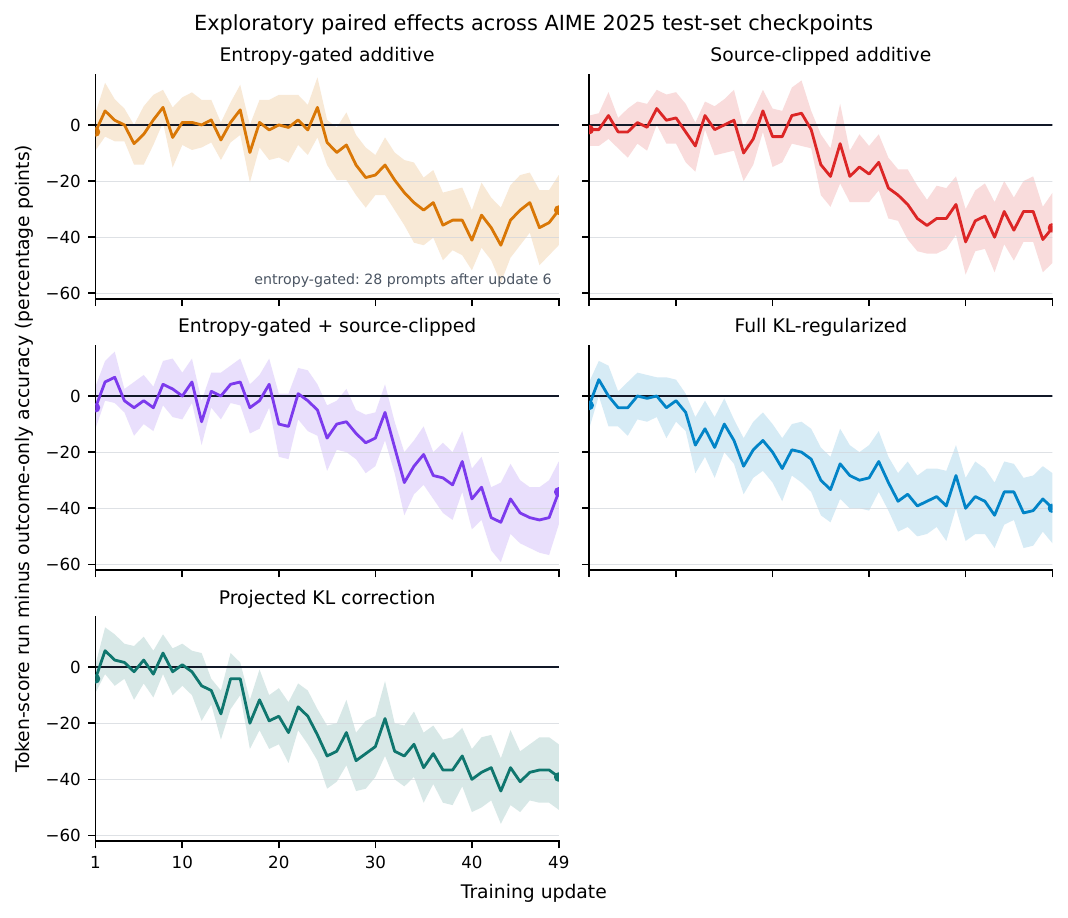}
\caption{\textbf{Prompt-paired effects at every eligible checkpoint.}
Each line is token-score configuration minus outcome-only accuracy, recomputed
on the same prompts at each update. Shading is a pointwise 95\%
prompt-bootstrap interval with 20,000 resamples. All comparisons use the same
AIME~2025 evaluation set. Update~49 is the fixed comparison.}
\label{fig:paired-effect-trajectories}
\end{figure}

We compare configurations at the latest checkpoint for which every
configuration has enough evaluation responses. This rule selects update~49
and excludes per-configuration best-checkpoint selection. All configurations
use the same AIME~2025 evaluation set, and each paired effect is recomputed on
\(\mathcal X_n\), the prompt intersection for that configuration and
outcome-only GRPO at update \(n\).
\Cref{fig:paired-effect-trajectories} reports these paired effects at every
eligible checkpoint.

\subsubsection{Equivalent-answer audit}

The evaluator first checks normalized exact equality and then symbolic
equivalence subject to domain restrictions. A numerical fallback uses fixed
points and precision chosen without configuration labels, while missing,
malformed, and truncated answers remain separate format outcomes.
High-consensus answers that disagree with the reference enter blinded
adjudication. Regrading changed seven rollout labels in the matched
evaluation.

\end{document}